\documentclass[a4paper,10pt]{article}
\pdfoutput=1
\usepackage{enumitem}
\setlist[enumerate]{leftmargin=.5in}
\setlist[itemize]{leftmargin=.5in}
\usepackage[top=1.25in, bottom=1.25in, left=1.25in, right=1.25in]{geometry}
\usepackage{amsfonts}
\usepackage{amsmath}
\usepackage{amsthm} 
\usepackage{amssymb} 
\usepackage{url}
\usepackage{bm}
\usepackage{bbm}
\usepackage{booktabs}
\usepackage{xcolor}
\usepackage{adjustbox}
\usepackage{mathtools} 
\usepackage{siunitx}
\usepackage{tikz-cd}
\tikzcdset{ampersand replacement=\&}
\usepackage{graphicx} 
\graphicspath{ {./pix} }
\usepackage{natbib}
\usepackage{float} 
\usepackage{algorithm} 
\usepackage{todonotes}

\usepackage{xcolor,algorithmicx,algorithm}
\usepackage{graphicx}
\usepackage{subcaption}
\DeclareMathOperator{\spn}{span}
\DeclareMathOperator{\Ima}{Im}

\DeclareMathOperator{\trace}{trace}
\DeclareMathOperator{\rank}{rank}

\DeclareMathOperator*{\argmin}{arg\,min \, }

\newcommand{\T}{{\top}}

\newcommand{\E}{{\mathbb E}}
\newcommand{\Pa}{{\mathbb P}}
\newcommand{\Q}{{\mathbb Q}}

\newcommand{\R}{{\mathbb R}}

\newcommand{\Ecal}{{\mathcal E}}

\newcommand{\Gcal}{{\mathcal G}}
\newcommand{\Hcal}{{\mathcal H}}
\newcommand{\Ical}{{\mathcal I}}

\newcommand{\Mcal}{{\mathcal M}}

\newcommand{\Ocal}{{\mathcal O}}
\newcommand{\Pcal}{{\mathcal P}}

\newcommand{\Rcal}{{\mathcal R}}

\newcommand{\Tcal}{{\mathcal T}}

\newcommand{\Xcal}{{\mathcal X}}
\newcommand{\Ycal}{{\mathcal Y}}
\newcommand{\Zcal}{{\mathcal Z}}

\newcommand{\bs}[1]{{\bm #1}}

\newcommand{\dd}{\operatorname{d}\!}

\newcommand{\isdef}{\mathrel{\mathrel{\mathop:}=}}

\newcommand{\StateIndent}{\hspace{\algorithmicindent}}

\newtheorem{proposition}{Proposition}[section]
\newtheorem{lemma}[proposition]{Lemma}
\newtheorem{theorem}[proposition]{Theorem}

\usepackage[hyperfootnotes=true,colorlinks,pagebackref=true,citecolor=blue]{hyperref}

\ifpdf
  \DeclareGraphicsExtensions{.eps,.pdf,.png,.jpg}
\else
  \DeclareGraphicsExtensions{.eps}
\fi

\title{Adaptive joint distribution learning\thanks{Michael Multerer gratefully acknowledges the Swiss National Science Foundation starting grant ``Multiresolution methods for unstructured data'' (TMSGI2\_211684). Paul Schneider gratefully acknowledges the Swiss National Science Foundation grant ``Scenarios'' (100018\_189086). We thank Rohan Sen, Marc Van Uffelen for helpful comments.}}
\author{Damir Filipovi\'c\footnote{EPFL and Swiss Finance Institute. Email: damir.filipovic@epfl.ch} \and Michael Multerer\footnote{Universit\`a della Svizzera italiana. Email: michael.multerer@usi.ch} \and Paul Schneider\footnote{Universit\`a della Svizzera italiana and Swiss Finance Institute. Email: paul.schneider@usi.ch}  }

\newcommand{\test}{\textnormal{test}}
\newcommand{\train}{\textnormal{train}}

\ifpdf
\hypersetup{
  pdftitle={Adaptive joint distribution learning},
  pdfauthor={D. Filipovi\'{c}, M. Multerer, and P.Schneider}
}
\fi

\begin{document} 

\maketitle

\begin{abstract}
We develop a new framework for estimating joint probability distributions using
tensor product reproducing kernel Hilbert spaces (RKHS). Our framework
accommodates a low-dimensional, normalized and positive model of a 
Radon--Nikodym derivative, which we estimate from sample sizes of up to several 
millions, alleviating the inherent limitations of RKHS modeling. 
Well-defined normalized and positive conditional distributions are natural 
by-products to our approach. Our proposal is fast to compute and accommodates 
learning problems ranging from prediction to classification. Our theoretical 
findings are supplemented by favorable numerical results.
\end{abstract}


\section{Introduction}
Most empirical applications in data science have a
formulation that depends on an unknown probability distribution. For
instance, linear regression can be viewed as an approximation of the conditional
expectation of the response variables given covariates, and classification
can be viewed as the task of minimizing the unknown probability of assigning a
wrong label to a variable. Both of these applications, and many others, have in
common that they approximate expectations with respect to the unknown, joint
stationary distribution through sample averages. If conditional
distributions were available on top of sample averages,  empirical researchers could assess marginal
effects and even complicated questions such as causality. However, there is no canonical model for conditional distributions based only on sample averages, and the literature
has accordingly developed nonparametric locally smoothed and sieve estimators
\citet{liracine06,grenander81}. An attractive alternative to the aforementioned models is the
embedding of conditional probability distributions in reproducing kernel Hilbert
spaces (RKHS). This technique has a long-standing tradition starting from 
covariance operators \citet{berlinetthomas-agnan04}, and kernel mean embedding
\citet{fukumizubachjorddan04,smolagrettonlescholkopf07,songetal09,
songfukumizugretton13}. It has been shown to be a highly effective machine
learning technique in finite samples, with improved properties over local
approaches used in extant nonparametric regressions
\citet{gruenewaelderetal12}. Traditional embedding of distributions in RKHS is
elegant and easily applied, and can be motivated from many different angles. For
instance, the work \citet{gruenewaelderetal12} connects conditional mean embedding
to the seminal vector-regression framework of \citet{micchellipontil05}.
In \citet{parkmuandet20}, the RKHS embedding is constructed from a
measure-theoretic point of  view, while \citet{songetal09} and
\citet{klebanovschustersullivan20} derive it from operator theory.

However, all of the above-mentioned approaches suffer from several conceptual problems.
They involve assumptions concerning inverse operators that are difficult to
test. Estimates of probabilities may become negative, and they are not
normalized, rendering the RKHS distribution embedding framework problematic in
many different situations of empirical work. A conditional covariance matrix
estimated with traditional RKHS distribution embedding, for instance, is not
guaranteed to be positive semidefinite, and for higher dimensions in particular
is unlikely to be. Likewise, widespread  classification loss functions such as
the logistic loss function, become difficult to handle in the face of
non-normalized and negative probabilities.

In this paper we propose a new framework  to estimate a  Radon--Nikodym derivative of the joint
distribution with respect to the product distribution of its marginals in the
spirit of \citet{nguyenwainwrightjordan10} and \citet{schusteretal20}. We term it
joint distribution learner (JDL). We construct JDL on a grid of sample points as
a normalized and positive joint distribution, such that its distance to the true
and unobserved population joint distribution can be decomposed additively into
two parts. The first vanishes as the number of data points grows. The second
constitutes the loss function that we provide in this paper, for which we also
derive a representer theorem. With a tensor product RKHS this yields an optimal
JDL that is bilinear in the feature maps pertaining to the RKHS used. From this
representer theorem, the loss function can be minimized through a simple
linear-quadratic optimization problem that can be processed through standard
off-the-shelf solvers, or proximal algorithms.

JDL comes  with several important advantages over the traditional embedding of
distributions in RKHS. First, it  enables us to represent, in particular,
constants and distributions of independent random variables, that are both problematic for the
traditional conditional mean embedding (CME) as shown by
\citet{klebanovschustersullivan20}. Second, JDL is normalized and positive by
definition, and thus satisfies the two fundamental structural properties of
probability measures. This is in contrast to the  traditional distribution
embedding, which is a mere projection of distributions that does not preserve
these properties. To the best of our knowledge, only \citet{muzellec2021note}
have considered these shape restrictions within embedding of distributions in
RKHS. From a positive and normalized joint distribution,  conditional
distributions can be computed in a straightforward way. Third, we construct our
Radon--Nikodym derivative directly in terms of a low-rank approximation
facilitated by the pivoted Cholesky decomposition from \citet{BJ02,BL77,FWAH+09,
HPS12}, that allows us to control the trade-off between precision and 
computational tractability with a sharply controlled a-posteriori error.
 
Our adaptive low-rank formulation alleviates the inherent scaling limitations of RKHS modeling by crucially leveraging the tensor RKHS hypothesis. It also
produces, at negligible additional computational cost, a basis transformation
that diagonalizes  the linear-quadratic optimization problem, making it amenable
to very large data sets. A specialization to finite-dimensional hypothesis
spaces, in particular polynomial RKHS with sample expectation inner products, 
automatically  produces this basis transformation, and yields pointwise
positivity constraints.

The principles behind our approach are generic, and applicable in particular 
also to CME. We thus provide a full description of the CME low-rank loss
function, and show that it benefits from the same diagonalization as JDL,
making it likewise easily solvable  in applications involving  several millions of
data points. In addition, we propose strategies to constrain the CME
optimization problem to account for normalization and positivity.

We provide numerical experiments with real and simulated data of up to
26 dimensions with up to ten million data points in prediction and
classification problems. The results  suggest that conditional distributions
obtained from  JDL perform favorably compared to CME, and even compared to more
specialized setups such as kernel logistic regression. They further show
computation times that allow both CME and JDL to be used in large-scale 
real-world applications.

The remainder of the paper is organized as follows. In
Section~\ref{sec_problem}, we formulate the estimation problem, establish
normalization and positivity as the requirements defining probability
distributions, and introduce an appropriate loss function. In 
Section~\ref{secRTHM}, we introduce our tensor product hypothesis space, and
prove a representer theorem subject to the structural constraints.
In Section~\ref{sec:lowrank}, we develop the generic low-rank framework, that we specialize to polynomials in Appendix \ref{sec:polyembedding}. 
In Appendix \ref{sec:conditionaldistributionembedding}, we
apply the low-rank framework to the conditional mean embedding, which is then
used to benchmark our method in extensive numerical experiments in 
Section~\ref{sec:experiments}. Section~\ref{sec:conclusion} concludes.
Appendix \ref{sec:klr} describes low-rank kernel logistic regression. For readability, all proofs are contained in Appendix \ref{sec:proofs}.
\section{Problem formulation}\label{sec_problem}
Let $X,Y$ be random variables taking values in some separable, complete metric
spaces $\Xcal$ and $\Ycal$, with joint population distribution $\Pa$ on the
product space $\Zcal\isdef\Xcal\times\Ycal$ endowed with a product metric
$d_\Zcal$. For instance, one might consider the 2-product metric
$d_\Zcal^2=d_\Xcal^2+d_\Ycal^2$. This minimal assumption assures the existence
of conditional distributions, see \citet[Theorem 10.2]{dud_02}, and the weak
convergence of the empirical sample distributions introduced below. In practical
examples, $\Xcal$ and $\Ycal$ are subsets of the Euclidean space.

Let $z_1=(x_1,y_1),\dots,z_{n}=(x_{n},y_{n})$ be a sample of points in $\Zcal$.
We aim at computing a probability measure $\Q$ supported on the grid
\(\Gcal\isdef\{(x_i,y_j): i,j=1,\dots,n\}\subset\Zcal\)
that approximates $\Pa$.
Such a measure $\Q$ can then be used to efficiently compute various
functionals, such as the conditional expectation $\E_\Q[ f(Y)|X=x]$ of a
function $f$ on $\Ycal$. Any such measure $\Q$ can be written as
\begin{equation}\label{Qhdef}
    \Q=\Q(h)= \frac{1}{n^2} \sum_{i,j=1}^n
    \big(1+h(x_i,y_j)\big)\delta_{x_i}\otimes\delta_{y_j}, 
\end{equation}
where \(\delta_x\) denotes the Dirac measure supported at \(x\) and
\(h\colon\Zcal\to\R\) is a function satisfying the normalization and
positivity (i.e., non-negativity) properties
\begin{align}
      \sum_{i,j=1}^n h(x_i,y_j) &=0,\label{norm}\\
  1+h(x_i,y_j) &\ge 0\quad \text{for all }(x_i,y_j)\in\Gcal.\label{posi}
\end{align}

In \eqref{Qhdef}, the Radon--Nikodym derivative model is decomposed into the sum of a constant prior of $1$ and the RKHS function $h$, for several reasons. Firstly, $h=0$ is a
valid choice satisfying the structural properties \eqref{norm} and \eqref{posi},
and thus makes the feasible set non-empty regardless of the chosen (linear)
hypothesis space.   Secondly,  $h=0$ also coincides with the  prior of independence. Below, when we estimate $h$ in a RKHS hypothesis space $\Hcal$  with regularization, shrinking $h$ adheres to the information-theoretic notion of minimizing mutual information.  Thirdly, omitting the constant term 1 in \eqref{Qhdef} would be problematic for commonly
used hypothesis spaces, such as Gaussian RKHS, that do not contain
constant functions.
In such cases, we can also interpret $1+h$ as an
element of the direct sum $\R\oplus\Hcal$, where the constant is fixed to 1. 
More generally, one may also consider an approach based on conditionally positive
definite kernels, see \citet{Wendland05}, which then also allows to optimize the
constant.

Accordingly, the marginal distributions of $\Q$ on $\Xcal$ and $\Ycal$ can be
computed as
\begin{equation}\label{QXYeq}
    \begin{aligned}
    \Q_X =\frac{1}{n^2} \sum_{i,j=1}^n
    \big(1+h(x_i,y_j)\big)\delta_{x_i},\quad
   \Q_Y = \frac{1}{n^2} \sum_{i,j=1}^n
    \big(1+h(x_i,y_j)\big)\delta_{y_j},
   \end{aligned}
\end{equation}
and its conditional distributions are given by
\begin{equation}\label{QXYYXcond}
    \begin{aligned}
    \Q_{X\mid Y=y} = \frac{ \sum_{i=1}^n (1+h(x_i,y))
    \delta_{x_i}}{\sum_{i=1}^n (1+h(x_i,y))},\quad
     \Q_{Y\mid X=x} = \frac{  \sum_{j=1}^n
    (1+h(x,y_j))\delta_{y_j}}{\sum_{j=1}^n (1+h(x,y_j))}.
   \end{aligned}
\end{equation}
Note that, as $h$ is defined on $\Zcal$, the measures $\Q_{X\mid Y=y}$ and
$\Q_{Y\mid X=x}$ naturally extend to all $x\in\Xcal$ and $y\in \Ycal$, albeit
they may become signed measures, and the denominators in \eqref{QXYYXcond} may
become zero and may cause numerical issues.

We denote by
\(
\hat \Pa\isdef\frac{1}{n}\sum_{i=1}^n \delta_{z_i}=
\frac 1 n\sum_{i=1}^n \delta_{x_i}\otimes\delta_{y_i}
\)
the empirical distribution with marginals
\(
\hat\Pa_X\isdef\frac 1 n\sum_{i=1}^n\delta_{x_i}\) and
\(\hat\Pa_Y\isdef\frac 1 n\sum_{i=1}^n \delta_{y_i}.
\)
The latter differ from $\Q_X$ and
$\Q_Y$ in general, as the following elementary lemma shows.
\begin{lemma}\label{lem:intro}
We have
\begin{enumerate}
\item $\Q_X=\hat\Pa_X$ if and only if $\sum_{i':x_{i'}=x_i}\sum_{j=1}^n 
h(x_{i'},y_j) = 0$
for all $i=1,\dots, n$.
\item $\Q_Y=\hat\Pa_Y$ if and only if $\sum_{j':y_{j'}=y_j}\sum_{i=1}^n 
h(x_i,y_{j'}) = 0$
for all $j=1,\dots, n$.
\end{enumerate}
\end{lemma}

To assess how far $\Q(h)$ is from the population distribution $\Pa$, we
decompose this distance into two parts, the distance between $\hat \Pa$ and
$\Pa$, and the distance between $\Q(h)$ and $\hat\Pa$. For the first part, we 
recall the definition of the dual-bounded Lipschitz metric measuring the
distance between two probability measures $P,Q$ on $\Zcal$ given by
\begin{equation}\label{betadef}
    \beta(P,Q)\isdef\sup_{\| f\|_{BL}\le 1  }\left |
    \int_\Zcal f \dd P - \int_\Zcal f \dd Q\right|.
\end{equation}
Herein, we set
\[
\|f\|_{BL}\isdef\sup_{z\in\Zcal}|f(z)|+\sup_{z_1\neq z_2}
\frac{|f(z_1)-f(z_2)|}{d_\Zcal(z_1,z_2)},
\] see \citet[Section 11.3]{dud_02} for the definitions.
In fact, $\beta$ metrizes weak convergence of probability measures on $\Zcal$. Therefore, $ \beta(\hat \Pa,\Pa)\to 0$ a.s.\
for $n\to\infty$ whenever $z_1,\dots,z_n$ is an i.i.d.\ sample drawn from $\Pa$,
see for instance \citet[Theorem 11.4.1]{dud_02}. Hence, for any $\epsilon_1>0$
there exists $n$ large enough such that\footnote{In the case of an absolutely continuous distribution $\Pa$ on $\Zcal=\R^d$, this may require $n$ of the order $\epsilon_1^{-d}$, see \citet[Proposition 2.1]{dud_69}.}
\begin{equation}\label{hatPaPaeps}
    \beta(\hat\Pa,\Pa)\le \epsilon_1.
\end{equation}

For the second part, we note that 
$\dd \Q(h)=(1+h)\dd \,(\hat \Pa_X \otimes \hat \Pa_Y)$
for any function $h$. This suggests that we measure the  distance between
$\Q(h)$ and $\hat \Pa $, similarly to \eqref{betadef}, by the squared
\emph{worst case loss}
\begin{align*}
 \Ecal(h)\isdef \sup_{\|f\|_{L^2_{\hat \Pa_X\otimes \hat \Pa_Y}}\le 1  }
 \bigg|\int_\Zcal f \dd\hat \Pa - \int_\Zcal f  \dd\Q(h)\bigg|^2 .
\end{align*}
Note that $\hat\Pa=\Q(\hat h)$ for any function $\hat h:\Zcal\to\R$ such that
\(
  1 + \hat{h}(x_i,y_j)=n\, \delta_{ij}.\)
Henceforth, we fix such a function $\hat h$. This implies that we can write
\begin{align*}
\Ecal(h)=  \sup_{\|f\|_{L^2_{\hat \Pa_X\otimes \hat \Pa_Y}}\le 1  }
\Big|\langle f,  \hat h-h\rangle_{L^2_{\hat \Pa_X\otimes \hat \Pa_Y}}
\Big|^2 =\big\| \hat h-h\big\|_{L^2_{\hat \Pa_X\otimes \hat \Pa_Y}}^2.
\end{align*}

Adding these two components, we arrive at the following result.
\begin{lemma}\label{lemdist}
Assume that \eqref{hatPaPaeps} holds and let $h$ be such that \eqref{norm} and
\eqref{posi} hold. Then the dual-bounded Lipschitz distance between $\Q(h)$ and
the population distribution $\Pa$ is bounded by
\(
    \beta(\Q(h),\Pa)\le \epsilon_1 + \Ecal(h)^{1/2}.
    \)
\end{lemma}

Lemma~\ref{lemdist} states that, for large enough $n$ such that
\eqref{hatPaPaeps} holds, it remains to find $h$ such that $\Ecal(h)$ is small,
so that $\Q(h)$ is close to the population distribution $\Pa$. In this
paper, we will choose $h$ optimally from a separable RKHS $\Hcal$ on $\Zcal$.
Specifically, we solve the convex optimization problem
\begin{equation}\label{optgen}
 \underset{h\in\Hcal \text{ s.t.\
 \eqref{norm}, \eqref{posi}}}{\text{minimize}}
 \big\{ \Ecal(h)+\lambda \|h\|_\Hcal^2\big\},
\end{equation}
for some regularization parameter $\lambda> 0$.

\section{Representer theorem}\label{secRTHM}
We assume a tensor product RKHS $\Hcal=\Hcal_\Xcal\otimes\Hcal_\Ycal$ as 
hypothesis space for the function \(h\colon\Zcal\to\R\), where $\Hcal_\Xcal$ and
$\Hcal_\Ycal$ are separable RKHS on \(\Xcal\) and \(\Ycal\), respectively. The
associated reproducing kernels are denoted by $k_\Xcal$ and $k_\Ycal$,
respectively. The reproducing kernel \(k\) of \(\Hcal\)
satisfies
\(
k\big((x,y),(x',y')\big)= k_\Xcal(x,x')k_\Ycal(y,y')\),
\(
x,x'\in\Xcal,\ y,y'\in\Ycal.
\)

In what follows, we  refer to the row vectors of canonical feature maps as
\begin{equation}\label{PhiXYdef}
    \begin{aligned}
{\bs\Phi}_X(\cdot)&\isdef[k_\Xcal(x_1,\cdot),\ldots, k_\Xcal(x_n,\cdot)],\quad
{\bs\Phi}_Y(\cdot)&\isdef[k_\Ycal(y_1,\cdot),\ldots, k_\Ycal(y_n,\cdot)],
\end{aligned}
\end{equation}
and to the associated kernel matrices by
\(
\bs K_X\isdef[{\bs\Phi}_X(x_i)]_{i=1}^n\),
\(\bs K_Y\isdef[{\bs\Phi}_Y(y_i)]_{i=1}^n.
\)

For any function $h$ on $\Zcal$ we define the matrix of function values
with respect to the grid \(\Gcal\) as
\(
    h|_{\Gcal}\isdef [h(x_j,y_i)]_{i,j=1}^n\in\R^{n\times n}.
\)
That is, rows and columns of $h|_{\Gcal}$ correspond to $y_i$ and $x_j$,
respectively, i.e., $( h|_{\Gcal})_{ij}=h(x_j,y_i)$. We denote by 
$\bm 1\in\R^n$ the column
vector consisting of 1's and by
\({\bs E}\isdef{\bm 1}{\bm 1}^\T\in\R^{n\times n}\) the matrix consisting of
1's.

The following representer theorem states that the solution to \eqref{optgen}
lies in the subspace spanned by the functions
$\bs\Phi_{X,i}\otimes\bs\Phi_{Y,j}$, $i,j=1,\dots, n$, in $\Hcal$.

\begin{theorem}\label{thm:representer}
    The solution to \eqref{optgen} is of the form
     \begin{equation}\label{eq:state_ij_id}
     h = h(\bs H)\isdef{\bs\Phi}_Y(\cdot){\bs H}
{\bs\Phi}_X(\cdot)^\T,
     \end{equation}
    for some $\bm H\in\R^{n\times n}$. That is, $h|_\Gcal
    = \bm K_Y \bm H \bm K_X$.
The linear constraints \eqref{norm} and \eqref{posi} accordingly read
\begin{align}
\bm 1^\T  \bm K_Y \bm H \bm K_X\bm 1 &=0,\label{eq:norma}\\
  \bm E+ \bm K_Y \bm H \bm K_X &\ge \bm 0,\text{ element-wise,}\label{posiC}
\end{align}
and problem \eqref{optgen} is equivalent to the $n^2$-dimensional optimization
problem
\begin{equation}\label{optgenC}
 \underset{\bm H\in\R^{n\times n} \text{ s.t.\
 \eqref{eq:norma}, \eqref{posiC}}}{\text{minimize}} \Rcal_\lambda(\bm H),
\end{equation}
where the regularized objective function is defined by
\begin{equation}\label{Rcaldef}
\begin{aligned}
    \Rcal_\lambda(\bm H)&\isdef \frac{1}{n^2}
    \trace{(\bs H^{\T}\bs K_Y\bs K_Y\bs H\bs K_X\bs K_X)}
 -\frac{2}{n} \trace{(\bs K_Y\bs H \bs K_X)}\\
 &\quad +\frac{2}{n^2}\trace{(\bs 1^{\T}\bs K_Y\bs H\bs K_X\bs 1 )}
 +\lambda \trace{(\bs H^{\T}\bs K_Y\bs H \bs K_X)}.
\end{aligned}
\end{equation}
\end{theorem}

With constraint \eqref{eq:norma} imposed, the third term on the right hand side
of \eqref{Rcaldef} is identically equal to zero. We nevertheless include it in
the sections below to consider also unconstrained versions of \eqref{optgenC},
without the constraints \eqref{eq:norma} and \eqref{posiC}.

\section{Adaptive low-rank approximation}\label{sec:lowrank}

Problem \eqref{optgenC} is of dimension $n^2$, which may become computationally 
too demanding for large sample sizes $n$. We therefore propose an efficient
low-rank representation of the kernel matrix that provides an approximation
of the full problem the error of which we quantify below in Lemma \ref{thmChol} together with \ref{lem:approx}. Specifically, we consider here an adaptive low-rank
approach based on the pivoted Cholesky decomposition for the kernel matrices
\(\bs K_X\) and \(\bs K_Y\).
We first present an algorithm that calculates a bi-orthogonal basis
transformation concurrently with the pivoted Cholesky decomposition in 
Section~\ref{sec:pivchol}. We then propose an additional rotation into a basis
that is orthogonal in both $\Hcal$ and $L^2_{\hat \Pa_X\otimes \hat \Pa_Y}$ in
Section \ref{sec:doubleorthogonal}. This double orthogonality  diagonalizes the
quadratic terms in  objective \eqref{Rcaldef}, greatly facilitating its
optimization.  
In Appendix \ref{sec:polyembedding}, we investigate the JDL
with a finite-dimensional polynomial basis, that, on top of its inherent low-rank nature, also allows for pointwise positivity through a sum-of-squares constraint.

\subsection{Pivoted Cholesky decomposition and bi-orthogonal basis}
\label{sec:pivchol}
In this section, we propose Algorithm~\ref{algo:bioChol} that computes an
incomplete Cholesky decomposition ${\bs L}_{m}{\bs L}_{m}^\T$ of any given
symmetric and positive semidefinite matrix $\bs K$, concurrently with a
bi-orthogonal basis transformation $\bs B_m$. The basis transformation is an
essential byproduct to efficiently solving the low-rank approximation of the
constrained optimization problem \eqref{optgenC} at large scale.

\begin{algorithm}[htb]\caption{Pivoted Cholesky decomposition}
\label{algo:bioChol}\
\begin{flushleft}
\begin{tabular}{ll}
\textbf{input:}  & symmetric and positive semidefinite
matrix ${\bs K}\in\mathbb{R}^{n\times n}$, 
tolerance \(\varepsilon>0\)\\
\textbf{output:} & low-rank approximation 
\({\bs L}_m\in\mathbb{R}^{n\times m}\)
with \({\bs K}\approx{\bs L}_{m}{\bs L}_{m}^\T\),\\
& matrix \({\bs B}_m\in\mathbb{R}^{n\times m}\)
                 with 
\({\bs B}_{m}^\T{\bs L}_{m}={\bs I}_{m\times m}\)
\end{tabular}
\end{flushleft}
\begin{algorithmic}[1]
\State Initialization: set $m\isdef 1$,
 ${\bs d}_0\isdef\operatorname{diag}({\bs K})$, \({\bs L}_0\isdef[\,]\),
 \({\bs B}_0\isdef[\,],\)
\State\textbf{while} \(\|{\bs d}_{m-1}\|_{1}>\varepsilon  \)
\State\StateIndent determine
$p_m\isdef \operatorname{arg}\max_{1\le i\le n}d_{m-1,i}$
\State\StateIndent compute \[{\bs\ell}_m
\isdef\frac{1}{\sqrt{d_{m-1,p_m}}}\big({\bs K}-{\bs L}_{m-1}
{\bs L}_{m-1}^\T\big){\bs e}_{p_m}\]
\State\StateIndent compute
\[{\bs b}_m
\isdef \frac{1}{\sqrt{d_{m-1,p_m}}}\big({\bs I}-{\bs B}_{m-1}
{\bs L}_{m-1}^\T\big){\bs e}_{p_m}\]
\State\StateIndent set \({\bs L}_m
\isdef [{\bs L}_{m-1},{\bs\ell}_m]\), \({\bs B}_m
\isdef [{\bs B}_{m-1},{\bs b}_m]\)
\State\StateIndent set \({\bs d}_m\isdef {\bs d}_{m-1}-{\bs\ell}_m
\odot{\bs\ell}_m\), where $\odot$ denotes the element-wise
product
\State\StateIndent set \(m\isdef m+1\)
\end{algorithmic}
\end{algorithm}

Based on the choice of the pivot element in
line 3 of Algorithm~\ref{algo:bioChol}, we obtain different
well known low-rank approximations of the kernel matrix.
The presented Algorithm~\ref{algo:bioChol} employs a greedy
strategy and always removes the
largest entry of the Schur complement, cp.\ \citet{HPS12}.
Alternatively, one may choose a random pivot element, resulting
in the well known Nystr\"om approximation, cp. \citet{Seeger}.
Finally, a maximal reduction of the trace is achieved by
selecting the pivot corresponding to the column of the Schur
complement with the largest Euclidean norm, cp. \citet{Higham}.
This comes at the cost of assembling the full kernel matrix, and quickly
becomes inefficient for larger sample sizes.
Hence, we opt here for the current version, since it is rather cost-efficient
and provides rigorous a posteriori error bounds for the approximation. To
decouple the matrix approximation error from its size, we shall consider a
relative error in our numerical experiments. That is, in line 2, we replace the
\emph{absolute} tolerance $\varepsilon$ by $\varepsilon \trace(\bs K)$, for a
\emph{relative} tolerance $\varepsilon$.

Employing the associativity of the matrix product in lines 4 and 5 of
Algorithm~\ref{algo:bioChol}, it is easy to see that the cost of computing
\({\bs\ell}_m\) and \({\bs b}_m\) is of order \(\Ocal(mn)\) each. Hence, the
cost of performing \(m\) steps of Algorithm~\ref{algo:bioChol} is of cost
\(\Ocal(m^2n)\). Moreover we emphasize that the full matrix \({\bs K}\) never
needs to be computed. It is rather sufficient to provide the pivot columns and
the diagonal. 

Validity of Algorithm~\ref{algo:bioChol} is substantiated by the following
theorem.
\begin{theorem}\label{thmChol}
For any tolerance \(\varepsilon>0\), Algorithm~\ref{algo:bioChol} computes
$n\times m$-matrices $\bm B_m$ and $\bm L_m$ with
\(m\leq\operatorname{rank}{\bs K}\) such that $\bm K - \bm L_m \bm L_m^\T$
is positive semidefinite and
\begin{align}
\operatorname{trace}\big({\bs K}-{\bs L}_m{\bs L}_m^\T\big)&
\leq\varepsilon,\label{Bprop0}\\
\Ima\bm B_m &= \spn\{{\bs e}_{p_1},\dots,{\bs e}_{p_m}\},\label{Bprop3}\\
\bm B_m^\T \bm L_m &= {\bm I}_{m\times m},\label{Bprop2}\\
\bm K \bm B_m &= \bm L_m\label{Bprop1}.
\end{align}
\end{theorem}

Below we employ Algorithm~\ref{algo:bioChol} to also speed up the computations of the conditional mean embedding considered in \citet{gruenewaelderetal12} resulting in a computational cost of \(\mathcal{O}(n m^2)\), a memory cost of \(\mathcal{O}(m^2)\), cp.\ \eqref{Bprop3}, and an evaluation cost of \(\mathcal{O}(m^2)\).  Alternatively, other established low-rank approaches known in
literature, such as \texttt{FALKON}, see \citet{meanti2020kernel},
or \texttt{EigenPro3}, see \citet{abedsoltan2023toward}, may be used.
By employing a matrix-free iterative approach, \texttt{FALKON} and 
\texttt{EigenPro3}
(with exact projection) achieve a computational cost of \(\mathcal{O}(nm)\) per
iteration. This is an improvement over the cost of 
\(\mathcal{O}(nm^2)\) of Algorithm~\ref{algo:bioChol}, if less than \(m\) iterations are required to achieve a satisfactory accuracy. However, the evaluation cost of CME in the matrix-free setting is 
\(\mathcal{O}(nm)\), which is larger than the \(\mathcal{O}(m^2)\), which we attain. Thus, matrix-free approaches such as \texttt{FALKON} or \texttt{EigenPro3} may be computationally favorable 
if only a few evaluations are required.

\subsection{Double-orthogonal basis}\label{sec:doubleorthogonal}
Building on the bi-orthogonal basis transformation $\bs B_m$, we can define an
additional rotation that diagonalizes the quadratic parts of the regularized
objective function \eqref{Rcaldef}, greatly simplifying the linearly constrained
quadratic optimization problem. From a spectral decomposition
$\bs V_m\bs \Lambda _m\bs V_m^{\T}=\bs L_m ^{\T}\bs L_m$ that is of cost
$\Ocal(m^3)$, we can compute
\begin{equation}\label{eq:spectralQ}
 \bs Q_m\isdef \bs B_m \bs V_m.
\end{equation}

Applying Algorithm~\ref{algo:bioChol} and
the additional rotation \eqref{eq:spectralQ} separately to both $\bs K_X$ and
$\bs K_Y$,  yields matrices $\bs Q _X$ and $\bs Q _Y$ of dimension $n \times m_X$
and $n \times m_Y$, respectively.
We then approximate the $n^2$-dimensional optimization problem \eqref{optgenC}
by parametrizing the argument through $\bs H = \bs Q_Y \tilde{\bs H}\bs Q_X^{\T}$
for some $m_Y\times m_X$-dimensional matrix $\tilde{\bs H}$. Inserting this
parametrization into $\Rcal_\lambda$, and using the identities
\begin{align*}
\trace\big(\bs H^{\T}\bs K_Y\bs H \bs K_X\big)
&=\trace\big(\tilde{\bs H}^{\T}\bs Q_Y^{\T}
\bs K_Y\bs Q_Y\tilde{\bs H} \bs Q_X^{\T}\bs K_X\bs Q_X\big)
=\trace\big(\tilde{\bs H}^{\T}\tilde{\bs H}\big), \\
\trace\big(\bs H^{\T}\bs K_Y\bs K_Y\bs H \bs K_X \bs K_X\big)
&=\trace\big(\tilde{\bs H}^{\T}\bs \Lambda _Y\tilde{\bs H}\bs \Lambda _X\big),
\end{align*}
gives the low-rank regularized objective function on $\R^{m_Y\times m_X}$ by
\begin{align*} 
    \tilde\Rcal_\lambda(\tilde{\bs H})\isdef \Rcal_\lambda
    (\bs Q_Y \tilde{\bs H}\bs Q_X^{\T})
    &=\frac{1}{n^2} \trace\big(\tilde{\bs H}^{\T}\bs \Lambda _Y\tilde{\bs H}
\bs \Lambda _X\big)-\frac{2}{n}\trace{\big(\bs L_Y\bs V_Y\tilde {\bs H}
\bs V_X^{\T}\bs L_X^{\T}\big)}\\
&\qquad +\frac{2}{n^2}\trace{\big(\bs 1^{\T}\bs L_Y\bs V_Y\tilde
{\bs H} \bs V_X^{\T}\bs L^{\T}_X\bs 1\big)} +\lambda \trace{(\tilde{\bs H}^{\T}
\tilde{\bs H})}.
\end{align*}
The linear constraints \eqref{eq:norma} and \eqref{posiC} read in terms of
$\tilde{\bs H}$ as
\begin{align}
\bs 1^{\T}\bs L_Y\bs V_Y\tilde {\bs H} \bs V_X^{\T}
\bs L_X^{\T}\bs 1 &=0,\label{eq:normatilde}\\
 \bs E+\bs L_Y\bs V_Y\tilde {\bs H} \bs V_X^{\T}
\bs L_X^{\T}&\ge \bm 0,\text{ element-wise, respectively.}\label{posiCtilde}
\end{align}
In summary, this leads to the low-rank optimization problem
\begin{equation}\label{optgenClowR}
 \underset{\tilde{\bm H}\in\R^{m_Y\times m_X} \text{ s.t.\
 \eqref{eq:normatilde}, \eqref{posiCtilde}}}{\text{minimize}}
 \tilde\Rcal_\lambda(\tilde{\bm H}).
\end{equation}
Combining this with \eqref{eq:state_ij_id}, the low-rank approximated solution
to \eqref{optgen}
is then given by
     \begin{equation}\label{eq:state_ij_idlowR}
     \tilde h =  {\bs\Psi}_Y(\cdot)\tilde{\bs H}
{\bs\Psi}_X(\cdot)^\T
     \end{equation}
for the rotated low-rank row vectors of basis functions
\(
{\bs\Psi}_X(\cdot)\isdef{\bs\Phi}_X(\cdot)\bs Q _X\)
and \({\bs\Psi}_Y(\cdot)\isdef{\bs\Phi}_Y(\cdot)\bs Q _Y.\)
That is, $\tilde h$ lies in the $m_X m_Y$-dimensional subspace
\begin{equation}\label{tilHdef}
    \tilde\Hcal \isdef  \spn \left\{\bs\Psi_{X,i}\otimes \bs\Psi_{Y,j} : 
    i=1,\dots,m_X,\, j=1,\dots,m_Y\right\}
\end{equation}
in $\Hcal$. As claimed above, the additional rotation forms
simultaneously orthogonal sets in $\Hcal_\Xcal$ and $L^2_{\hat{\mathbb{P}}_X}$,
and $\Hcal_\Ycal$
and $L^2_{\hat{\mathbb{P}}_Y}$, respectively. This can be seen from the
following identities,
\begin{equation}\label{orthoPsiH}
\left [\langle{\bs\Psi}_{X,i},{\bs\Psi}_{X,j}
\rangle_{\Hcal_\Xcal}\right ]_{i,j=1}^{m_X}={\bs I},
\quad
\left [\langle{\bs\Psi}_{Y,i},{\bs\Psi}_{Y,j}
\rangle_{\Hcal_\Ycal}\right ]_{i,j=1}^{m_Y}={\bs I}
\end{equation}
and
\[\left [\langle{\bs\Psi}_{X,i},{\bs\Psi}_{X,j}
\rangle_{L^2_{\hat{\mathbb{P}}_X}}\right ]_{i,j=1}^{m_X}
=\frac{1}{n}{\bs\Lambda}_X,
\quad
\left [\langle{\bs\Psi}_{Y,i},{\bs\Psi}_{Y,j}
\rangle_{L^2_{\hat{\mathbb{P}}_Y}}\right ]_{i,j=1}^{m_Y}
=\frac{1}{n}{\bs\Lambda}_Y,\]
for the two diagonal eigenvalue matrices $\bs \Lambda _X$ and $\bs \Lambda _Y$.

\subsection{Positivity constraint tightenings}
In this section, we discuss the positivity constraint \eqref{posiC}, and
describe two different tightenings. The first, computationally intensive,
tightening is built upon bounded kernels, and ensures pointwise positivity. 
The second is a computationally attractive single-inequality tightening of 
\eqref{posiC}. The choice between these two types of positivity models,
pointwise versus data-almost-sure  can be motivated through weighing the 
trade off between empirical and computational needs and capacities. 
In the sections below, we exclusively use the second kind, that we find to
be empirically attractive with only very few constraint violations 
out-of-sample. In Appendix \ref{sec:polyembedding}, we furthermore describe
pointwise positivity through a sum-of-squares constraint.
We suggest to use pointwise positivity when the number of 
data points is small and the probability to evaluate far away from the 
observed data points is large.
\subsubsection{Pointwise positivity for bounded kernels}
Assuming bounded kernels on both $\Xcal$ and $\Ycal$, we have that 
\(\phi_{X,j}(z)\phi_{Y,i}(z)\in[a_{i,j},b_{i,j}]\), for
\(z\in\Zcal\), for some finite \(a_{i,j}<b_{i,j}\), for all 
\(i=1,\ldots,m_Y, j=1,\ldots, m_X\). Accordingly, we define the hyperrectangle
\(C\isdef\prod_{i=1,\ldots,m_Y; j=1, \ldots, m_X }[a_{i,j},b_{i,j}]\) and the
convex set 
\(\mathcal C\isdef\big\{ \bm H \in \mathbb R ^{m_Y\times m_X}\mid 
\min _{{\bm x}\in C}{\bm x}^{\intercal}\operatorname{vec}(\bm H)+1\geq 0\big\},
\)
where $\operatorname{vec}(\cdot)$ denotes the operator that concatenates the 
columns of a matrix into a column vector. Clearly, for any $\bm H\in \mathcal C$
we have $1+h(\bm H)\ge 0$ pointwise in $\Zcal$, which implies \eqref{posi}, or
equivalently \eqref{posiC}. Decomposing $\operatorname{vec}(\bm H)
=\bm h ^{+}-\bm h ^{-}$, with $\bm h^{-},\bm h^{+}\succeq \bm 0$, we derive the
following equivalent characterization of $\mathcal C$.
\begin{lemma}[Non-negativity for bounded kernels]\label{lem:positive}
Set $\bm a\isdef 
\operatorname{vec}([a_{i,j}]_{i=1,\ldots, m_Y, j=1,\ldots, m_X})$ and 
$\bm b\isdef \operatorname{vec}([b_{i,j}]_{i=1,\ldots, m_Y, j=1,\ldots, m_X})$. 
We have that $ \bm H\in \mathcal C$ if and only if 
 \begin{equation}\label{eq:condpos}
  \bm a^{\intercal}\bm h^{+}-\bm b ^{\intercal} \bm h^{-}+1 \geq 0.
 \end{equation}
 In this case, \eqref{posiC} is satisfied.
\end{lemma}

The above pointwise tightening could also be applied to the positivity 
constraint~\eqref{posiCtilde} expressed in the rotated basis functions.

\subsubsection{Single-inequality tightening}
The positivity constraint \eqref{posiCtilde}, and \eqref{posiC} and its 
pointwise tightening \eqref{eq:condpos}, are sets of $n\times n$ inequalities.
For large $n$, as in our experiments below, this may become computationally
prohibitive. We thus replace it with a single-inequality tightening based on the
maxima and minima of the columns of the matrices $\bs L_X\bs V_X$ and
$\bs L_Y\bs V_Y$, respectively, i.e.,
\begin{align*}
\overline{\bs c}_{X,j}\isdef
\max _{i=1,\ldots,n}[\bs L_X\bs V_X]_{i,j},
\quad & \underline{\bs c}_{X,j}
\isdef \min _{i=1,\ldots,n}[\bs L_X\bs V_X]_{i,j},\quad j=1,\ldots,m_X,\\
 \overline{\bs c}_{Y,j}
 \isdef \max _{i=1,\ldots,n}[\bs L_Y\bs V_Y]_{i,j},
 \quad  & \underline{\bs c}_{Y,j}
 \isdef \min _{i=1,\ldots,n}[\bs L_Y\bs V_Y]_{i,j},\quad j=1,\ldots,m_Y.
\end{align*}

For any real number $a$, we denote by $a^+\isdef\max\{a,0\}$ and 
$a^-\isdef-\min\{a,0\}$ its positive and negative parts, respectively. Define
the $m_Y\times m_X$ matrices $\underline{\bs C}$ and $\overline{\bs C}$ by
\begin{align}
\underline{\bs C}_{i,j} &\isdef   (\underline{\bs c}_{Y,i})^+
(\underline{\bs c}_{X,j})^+ + (\overline{\bs c}_{Y,i})^-
(\overline{\bs c}_{X,j})^- -\max\left\{ (\overline{\bs c}_{Y,i})^+
(\underline{\bs c}_{X,j})^-, (\underline{\bs c}_{Y,i})^- 
(\overline{\bs c}_{X,j})^+\right\},\label{minCdef}\\
\overline{\bs C}_{i,j} &\isdef \max\left\{ (\overline{\bs c}_{Y,i})^+ 
(\overline{\bs c}_{X,j})^+, (\underline{\bs c}_{Y,i})^- 
(\underline{\bs c}_{X,j})^-\right\} -  (\overline{\bs c}_{Y,i})^- 
(\underline{\bs c}_{X,j})^+ - (\underline{\bs c}_{Y,i})^+ 
(\overline{\bs c}_{X,j})^-  .\label{maxCdef}
\end{align}
Then replace $n\times n$ positivity constraints \eqref{posiCtilde} by the single
constraint
\begin{equation}\label{eq:tightening}
    1+\trace\big(\underline{\bs C}^\T \tilde {\bs H}^+\big)
    -\trace\big(\overline{\bs C}^\T \tilde {\bs H}^-\big)\ge 0,
\end{equation}
where we define the $m_Y\times m_X$ matrices $\tilde{\bs H}^+$ and 
$\tilde{\bs H}^-$ to consist of the positive parts ${\tilde H_{i,j}}^+$ and
negative parts ${\tilde H_{i,j}}^-$, respectively. The following lemma asserts
that this is a tightening indeed.

\begin{lemma}\label{lemposi}
If $\tilde {\bs H} $ satisfies \eqref{eq:tightening} then it also satisfies 
\eqref{posiCtilde}.
\end{lemma}

Evidently, the cost of computing \eqref{minCdef} and \eqref{maxCdef} is much
lower than for the right hand side of \eqref{minCeq} and \eqref{maxCeq}, which
requires $\Ocal(n^2)$, in terms of given matrices $\bs L_Y\bs V_Y$ and 
$\bs L_X\bs V_X$.

\subsection{Elementary approximation error bounds}

Note that the low-rank approximated solution $\tilde h\in\tilde\Hcal$ in 
\eqref{eq:state_ij_idlowR} is different from the orthogonal projection in 
$\Hcal$, say $h_{\tilde\Hcal}$, of the solution $h$ in \eqref{eq:state_ij_id} on
the subspace $\tilde\Hcal$ in \eqref{tilHdef}. The next result gives elementary
lower and upper bounds on the low-rank approximation error, 
$\|h-\tilde h \|^2_{\Hcal}$, in terms of $h_{\tilde\Hcal}$ and the tolerance 
$\varepsilon$. This illustrates the trade-off between computational tractability
and precision. We denote by $\|\cdot\|_F$ the Frobenius norm.
\begin{lemma}\label{lem:approx}
Let  $h$ from \eqref{eq:state_ij_id} and $\tilde h$ from
\eqref{eq:state_ij_idlowR} denote the solutions to \eqref{optgen} and
\eqref{optgenClowR}, respectively, solved with low-rank tolerances 
\(\trace\big({\bs K}_X-{\bs L}_X{\bs L}_X^\T\big)<\varepsilon\)
and \(\trace\big({\bs K}_Y-{\bs L}_Y{\bs L}_Y^\T\big)<\varepsilon\). Then the
orthogonal projection of $h$ on $\tilde\Hcal$ is of the form
\(h_{\tilde\Hcal}= \bs\Psi_Y (\cdot) \bm H_{\tilde\Hcal} 
\bs\Psi _X^{\T} (\cdot)\) with coefficient matrix given by
\begin{equation}\label{eq:orthoproj}
 \bm H_{\tilde\Hcal} = \argmin_{\bs H'\in 
 \R ^{m_Y\times m_X}} {\|h-{\bs\Psi}_Y(\cdot)\bs H' 
{\bs\Psi}_X(\cdot)^\T}\|_{\Hcal}^2=\bm V_{Y}^{\T}
\bm L_{Y}^{\T}\bm H\bm L_{X}\bm V_{X}.
\end{equation}
Moreover, we have
\[  \| \bm H_{\tilde\Hcal}-\tilde{\bm H} \|_F^2\leq \|h-\tilde h \|^2_{\Hcal}
 \leq \varepsilon  \, \|{\bs H}\|_F^2 \big(\trace{\bs K}_X+\trace{\bs K}_Y) 
   +  \| \bm H_{\tilde\Hcal}-\tilde{\bm H} \|_F^2.\]
%
\end{lemma}

\section{Numerical experiments}\label{sec:experiments}
To assess the performance of the proposed JDL, we conduct numerical prediction 
and classification experiments with both real and simulated data. We consider
conditional second-moment matrices, as well as binary classification as test
tasks. Both types of numerical experiments crucially rely upon positivity and
normalization of the implied probability measures, as any conditional second
moment matrix ought to be positive semidefinite, and the classification
probabilities  ought to take values in the unit interval. These are testable
structural features, that we evaluate for different sample sizes and dimensions,
along with suitable loss functions. As a natural benchmark for our proposed 
JDL, we consider \emph{conditional mean embedding} (CME) proposed by 
\citet{songetal09} and \citet{gruenewaelderetal12}, as well as kernel logistic
regression (KLR) for binary classification.

CME's construction also facilitates a tensor product RKHS, 
$\Hcal=\Hcal _X \otimes \Hcal _Y$. However, through convenient cancellations,
see Footnote \ref{foot:lucky}, $\Hcal _Y$ does not explicitly feature other than
as a nuisance space that is merely used for  function evaluation in its original
definition. In particular, the kernel matrix $\bs K_Y$ never needs to be
computed or used in any way.  However, for large sample sizes CME becomes
computationally intractable.

To put  conditional  distribution embedding on comparable computational footing
with JDL proposed in this paper, we develop  a   low-rank version for CME 
in Appendix~\ref{sec:conditionaldistributionembedding}, along with the
structural constraints. For KLR we develop the corresponding low-rank
representation in Appendix \ref{sec:klr}.
Low-rank CME is computationally tractable also for large-scale problems, but it
comes at the price of introducing  a low-rank version of $\bs K_Y$ that is not
present in the full unconstrained $n\times n$ CME problem. This leaves us with
the choice between the necessity of validating only one kernel parameter, but
not being able to process large data sets, or having to validate two kernel
parameters for large data sets. In our numerical experiments we opt for the
latter in order to process  large $n$  data sets, as already $n=1000$ poses a
formidable matrix inversion problem.

We use two types of hypothesis spaces for JDL, a Gaussian RKHS, to which we
refer to as JDL, and a polynomial RKHS with order $q=4$, as described in
Section~\ref{sec:polyembedding}, referred to as JPDL. CME and KLR are
implemented with Gaussian RKHS.  For all experiments, we implement
Algorithm~\ref{algo:bioChol} (Theorem~\ref{thmChol}) with absolute tolerance
$\varepsilon$ in line 2 (in \eqref{Bprop0}) replaced by $\varepsilon
\trace(\bs K)$, for a relative tolerance $\varepsilon$.

We consider both structurally constrained, and unconstrained versions of JDL,
JPDL, and CME. For JDL the constraints are incorporated through
\eqref{eq:normatilde} and \eqref{eq:tightening}, for JPDL through
\eqref{eq:normatilde} and  \eqref{eq:semdefcone}. CME is less suitable to
normalization, as \eqref{normCMEfull} shows that it amounts to  $n$ restrictions
even in the low-rank case. Clearly, this constraint is impractical for large
$n$, and infeasible in general, unless the dimension $m_X m_Y$ of the argument
is greater than $n$. We therefore only use the weaker constraint demanding
unconditional normalization \eqref{normCME}. Positivity for CME, described in
\eqref{posCME}, could in principle be imposed through the same one-equation
tightening \eqref{eq:tightening} as for JDL. However, this constraint turns out
to be infeasible for all examples and parameterizations developed in this
section. We therefore use the weaker constraint of unconditional positivity from
\eqref{posCMEave}, that is feasible for most parameterizations and examples 
below.

\subsection{Conditional  moment matrices}
Simulated data may reveal the behaviour of the embeddings with respect to the
sample size, as well as the low-rank approximation in a controlled environment.
We consider two types of distributions, Gaussian and uniform, as
conditional expectations are readily available for both.

To obtain draws from multivariate Gaussian distributions with varying degrees of
dependency, we first generate $j=1,\ldots, 100$ random correlation matrices of
dimension $2d$, where $d=3$, following the algorithm described in 
\citet{ilyahensen21}. Assuming zero means and unit standard deviations for
simplicity, we subsequently sample from the corresponding multivariate Gaussian
distribution data sets $z_1^{(j)},\ldots, z_n^{(j)}$ of size 
$n=10^3, 10^4,10^5,10^6,10^7$, for each of the $j=1,\ldots, 100$ correlation
matrices. We split each of the data sets into a training set of size 
$n_{\train}= 4n/5$ and a validation set of size $n/5$ to validate the kernel and
regularization hyperparameters. To test the distributions out-of-sample, we draw
for each of the $j=1,\ldots, 100$ correlation matrices additional
$n_{\test}=5{,}000$ data points from the same distribution that generated the
training and validation sets. We additionally investigate very
small data sets of size $n=50,100, 200$ with respect to structural behavior of
the moment matrices,  where each of the realizations $z_i^{(j)}=(u_{i,1}^{(j)},
(u_{i,1}^{(j)}+u_{i,2}^{(j)})/2,\ldots, (u_{i,1}^{(j)}+\cdots 
+ u_{i,2d}^{(j)})/(2d))$, with $u_{i,1}^{(j)},\ldots, u_{i,2d}^{(j)}$ i.i.d.\ 
draws from a uniform distribution on the unit interval.

Second-moment matrices play an important role in many applications involving
regression-type models and quadratic forms of optimization problems that are
convex only if they are positive semidefinite. If they are generated by
probability measures, they are always positive semidefinite. Consequently, a
non positive semidefinite second-moment matrix indicates that it has not been
generated by a bona fide probability measure.  Ubiquitous across many fields,
second-moment matrices thus constitute an important test case. We therefore keep
track of the properties of the resulting conditional moment matrices for each of
the simulations, on top of how well they predict squares of random variables.

Corresponding to the task of estimating conditional second-moment matrices, we
introduce the test function $\bs \Tcal : \Ycal \mapsto \R ^{d\times d}$,
\( \bs \Tcal(y)\isdef [y_1,\ldots, y_d] ^{\T} [y_1,\ldots, y_d].\)

Writing  $\Mcal$ as placeholder for the models JDL (joint distribution learning,
Gaussian RKHS), JDPL (joint distribution learning, polynomial RKHS), CME 
(conditional mean embedding, Gaussian RKHS),   we use the second-moment squared 
loss function
\begin{equation}\label{eq:secmomloss}
 \frac{1}{n_{\test}}\sum _{i=1}^{n_{\test}} \frac{\Big\|  
 \mathbb E^{\mathbb Q_{Y|X=x^{\test,j}_i}}[ \bs \Tcal(Y)]-\mathbb E^{\mathbb
Q_{Y|X=x^{\test,j}_i}}_{\bs \theta _j, \Mcal}[\bs \Tcal(Y)]\Big\|^2_{F}}{\Big\|
\mathbb
E^{\mathbb Q_{Y|X=x^{\test,j}_i}}[ \bs \Tcal(Y)]\Big\|_{F}^2}
\end{equation}
to (validate) test the models, where $
x^{\test,j}_{1},\dots, x^{\test,j}_{n_{\test}}$ denotes the test sample (for the
validation this is replaced by the validation sample), and $\bs \theta _j$ the
(to be) validated hyperparameters pertaining to model $\Mcal$, in experiment $j$.

\begin{figure}[htb]
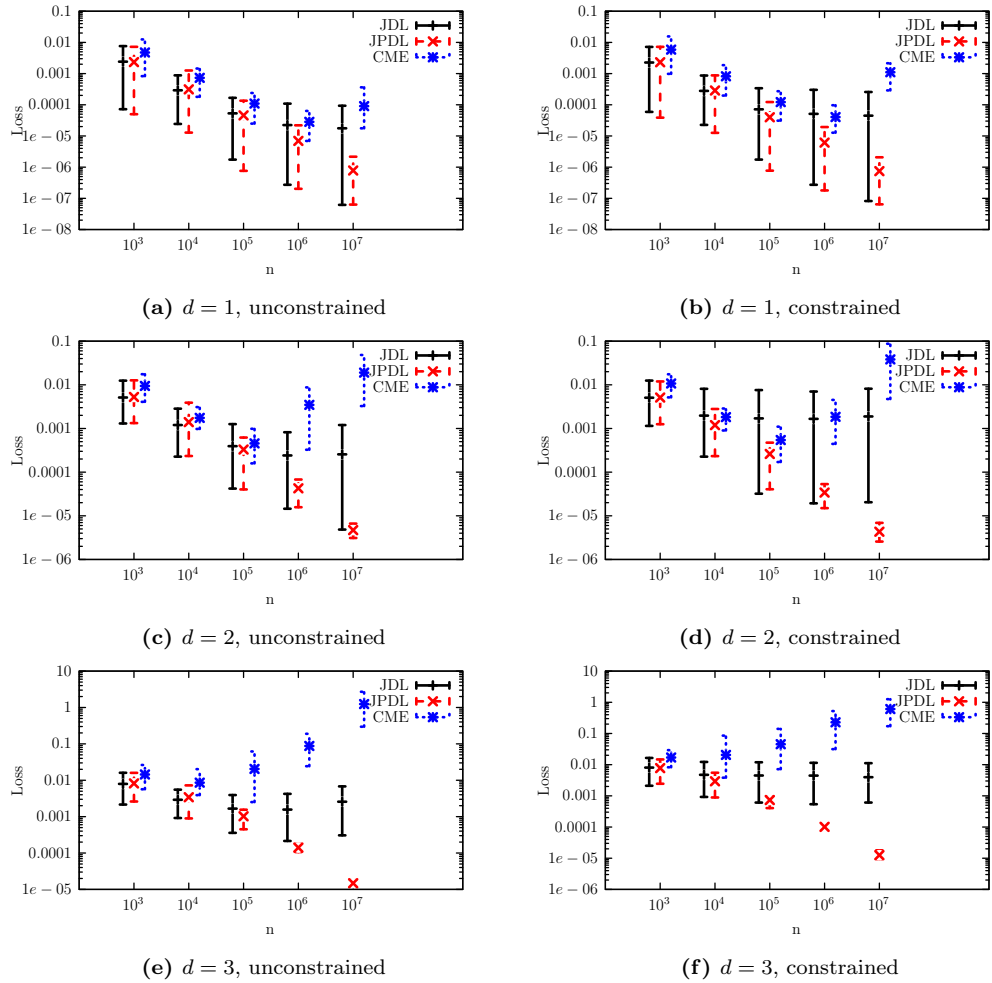

 \begin{subfigure}{0.49\textwidth}
 \scalebox{0.55}{\input{figs/IN_fig_POL_3}}
 \caption{\label{fig:uccovarloss_10}$d=3$, unconstrained}
 \end{subfigure}
  \begin{subfigure}{0.49\textwidth}
 \scalebox{0.55}{\input{figs/IN_fig_POL_C_3}}
 \caption{\label{fig:covarloss_10}$d=3$, constrained}
 \end{subfigure}
  \caption{\label{fig:covarloss}Second-moment squared loss. The panels show
  loss function \eqref{eq:secmomloss} evaluated over $n_{\test}=5{,}000$ samples
for the   joint distribution learner (JDL), the polynomial joint distribution
learner (JPDL), and  the conditional mean embedding (CME)  on the $y$-axis.
The $x$-axis shows the number $n$ of data points used for training and
validation. The data are generated from a mean-zero, unit standard deviation
multivariate Gaussian distribution with covariance matrix sampled from the
algorithm proposed by \citet{ilyahensen21}. For $n\leq 10^5$, low-rank algorithm
\ref{algo:bioChol} is applied to the kernel matrices with tolerance
$\varepsilon=10^{-3}$, for $n= 10^6$ with $\varepsilon=10^{-2}$, and for
$n=10^7$ with $\varepsilon=10^{-1}$. JPDL features $\varepsilon=0$. }
\end{figure}

We compare accuracy, structural preservation, and speed of the methods. For each
of the criteria, we compare the mean and $5\%$ and $95\%$ quantiles of the
sampling distribution of the second-moment squared loss \eqref{eq:secmomloss}
over the $j=1,\ldots, 100$ runs of the competing   models, considering both the
unconstrained and the constrained versions. We consider unconstrained versions
to gauge how much the structural restrictions tilt the embeddings away from the
statistical optimum. Unconstrained problems are solved on two, the constrained
problems on ten Intel Xeon 2.3 GHz processors. For $n\leq 10^5$, low-rank
algorithm \ref{algo:bioChol} is applied to the kernel matrices with relative
tolerance $\varepsilon=10^{-3}$, for $n= 10^6$ with $\varepsilon=10^{-2}$, and
for $n=10^7$ with $\varepsilon=10^{-1}$. JPDL features  $\varepsilon=0$ by
construction. Larger relative tolerances for larger $n$ are chosen to keep the
computation times and memory requirements low also for large data sets. In the
one-dimensional case, we furthermore investigate the behaviour of the embeddings
with varying relative tolerance $\varepsilon$ when  $n$ grows large.

\begin{figure}[htb]
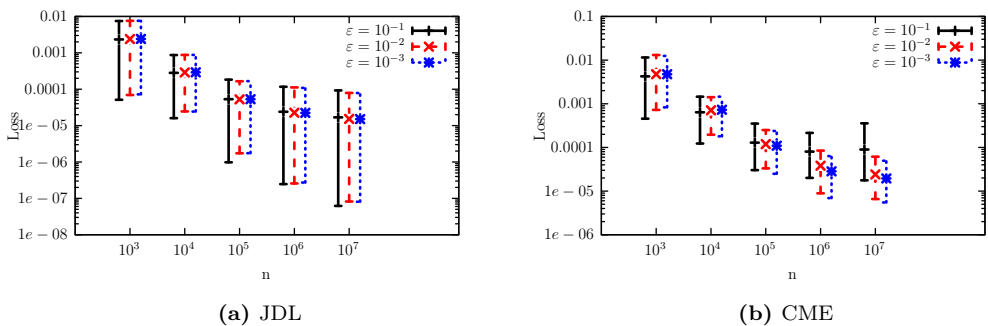

 \begin{subfigure}{0.49\textwidth}
 \scalebox{0.55}{\input{figs/IN_fig_JDL_eps}}
 \caption{\label{fig:JDLtol}JDL}
 \end{subfigure}
 \begin{subfigure}{0.49\textwidth}
 \scalebox{0.55}{\input{figs/IN_fig_CME_eps}}
 \caption{\label{fig:CMEtol}CME}
 \end{subfigure}\\
  \caption{\label{fig:tolerance}Low-rank tolerance and data size. The panels
  show loss function \eqref{eq:secmomloss} evaluated over $n_{\test}=5{,}000$
samples for the joint distribution learner (JDL)  and  the conditional mean
embedding (CME) on the $y$-axis. The $x$-axis shows the number $n$ of data
points used for training and validation. The data are generated from a
mean-zero, unit standard deviation multivariate Gaussian distribution with
covariance matrix sampled from the algorithm proposed by \citet{ilyahensen21}.
Tolerances are indicated in the figures.}
\end{figure}

Figure \ref{fig:covarloss} shows the second-moment squared loss
\eqref{eq:secmomloss} as a function of the data sample size $n$ for $d=1,2,3$.
JPDL dominates JDL, wich again dominates CME. All models except JPDL may
incur losses with increasing in $n$, which is due to the increasing relative
tolerance level $\varepsilon$, and is most pronounced for CME. Indeed, Figure
\ref{fig:tolerance} illustrates the trade-off between $n$ and $\varepsilon$ for
CME in the one-dimensional case. Panel \ref{fig:CMEtol} shows that losses may
become larger with increasing $n$, if also $\varepsilon$ is increased. At least
in the one-dimensional case this is not so for JDL: Panel \ref{fig:JDLtol}
indicates it to be largely robust to changes in the tolerance. Figure
\ref{fig:covarloss} further indicates that CME losses are particularly
pronounced for higher dimensions.

\begin{figure}[htb]
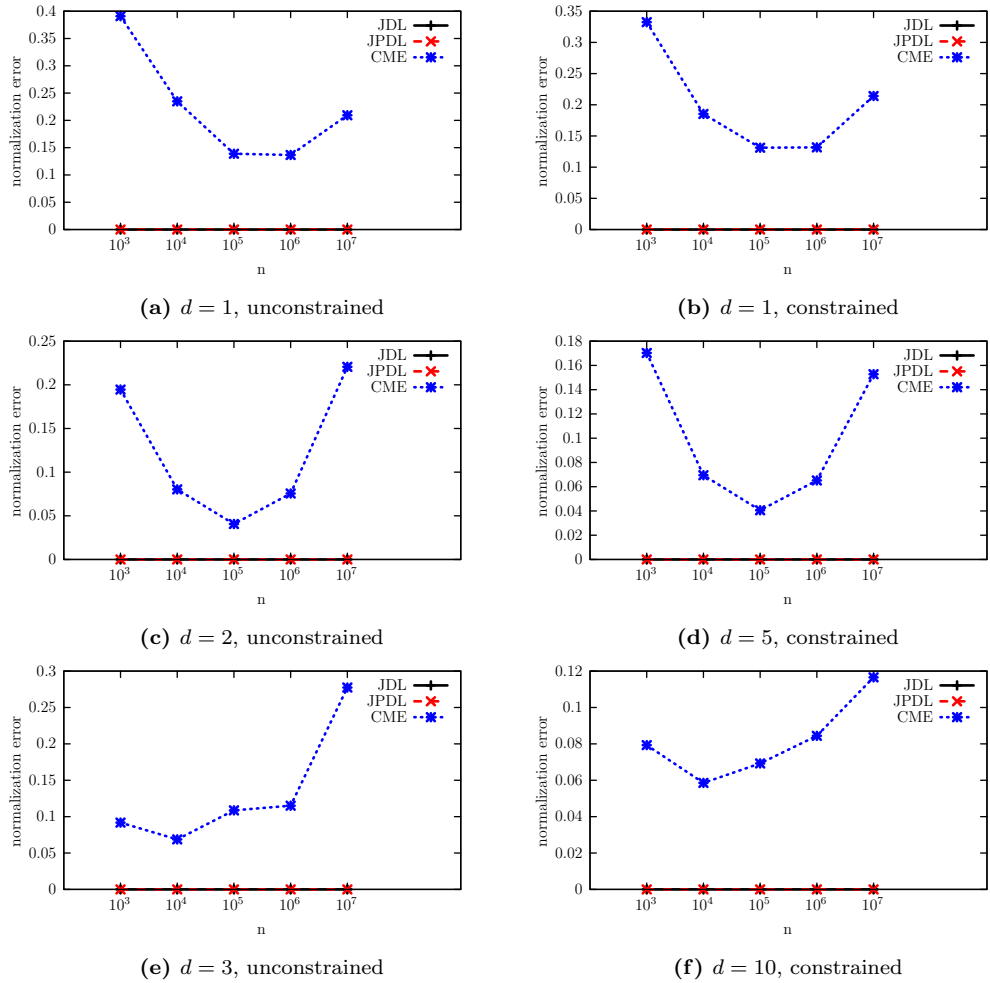

 \begin{subfigure}{0.49\textwidth}
 \scalebox{0.55}{\input{figs/IN_fig_POL_dev_3}}
 \caption{\label{fig:uccovardev_10}$d=3$, unconstrained}
 \end{subfigure}
  \begin{subfigure}{0.49\textwidth}
 \scalebox{0.55}{\input{figs/IN_fig_POL_C_dev_3}}
  \caption{\label{fig:covardev_10}$d=3$, constrained}
 \end{subfigure}
 \caption{\label{fig:covardev}Normalization error. The panels show the maximal
 normalization error over $n_{\test}=5{,}000$ samples for the unconstrained 
joint distribution learner (JDL), the polynomial joint distribution learner
(JPDL), and the conditional mean embedding (CME) on the $y$-axis. The $x$-axis
shows the number $n$ of data points used for training and validation. The data
are generated from a mean-zero, unit standard deviation multivariate Gaussian
distribution with covariance matrix sampled from the algorithm proposed by
\citet{ilyahensen21}. For $n\leq 10^5$, low-rank algorithm \ref{algo:bioChol}
is applied to the kernel matrices with tolerance $\varepsilon=10^{-3}$,
for $n= 10^6$ with $\varepsilon=10^{-2}$, and for $n=10^7$
with $\varepsilon=10^{-1}$. JPDL features  $\varepsilon=0$.}
\end{figure}

\begin{figure}[htb]
\begin{center}
  \scalebox{0.55}{\input{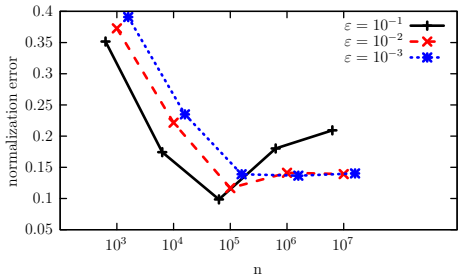}}
 \end{center}
  \caption{\label{fig:cmenormerror}CME normalization error, low-rank and data
  size trade-off. The panel shows the maximal normalization error over
$n_{\test}=5{,}000$ samples for   the conditional mean embedding (CME). The 
$x$-axis shows the number $n$ of data points used for training and validation. 
The data are generated from a mean-zero, unit standard deviation multivariate
Gaussian distribution for $d=3$, with covariance matrix sampled from the algorithm proposed
by \citet{ilyahensen21}. Tolerances are indicated in the legend.}
\end{figure}

Figure \ref{fig:covardev} shows the worst-case normalization deviation,
\[\max _{i=1,\ldots, n_{\test};\, j=1,\ldots, 100}\Big| \mathbb E^{\mathbb
Q_{Y|X=x^{\test,j}_i}}_{\bs \theta _j, \Mcal}[1]-1\Big|,\]
over all experiments. By construction, JDE and JPDE are normalized also without
constraints. On the contrary,  CME  exhibits sizable normalization errors that
are inversely related to sample size, and comove with the precision of the
low-rank approximation. Figure \ref{fig:cmenormerror} illustrates this pattern
in one dimension: at a relative tolerance of $\varepsilon = 10^{-3}$ the
worst-case normalization error decreases monotonically with $n$. For both
$\varepsilon = 10^{-2}$ and $\varepsilon =10^{-1}$ the normalization error is
lowest at $n=10^5$ data points, but then increases only slightly with $n$ for
$\varepsilon = 10^{-2}$, while it grows  quite substantially for
$\varepsilon = 10^{-1}$. Figure \ref{fig:covardev} shows that structural
constraints mitigate normalization errors.

\begin{figure}[htb]
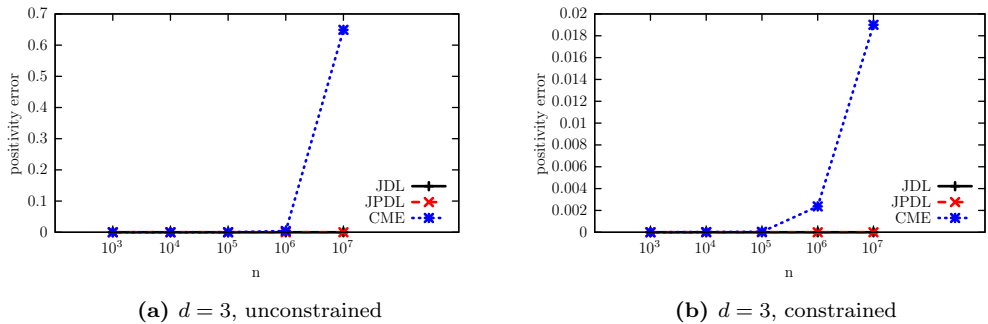

 \begin{subfigure}{0.49\textwidth}
 \scalebox{0.55}{\input{figs/IN_fig_POL_neg_3}}
 \caption{\label{fig:uccovarpos_10}$d=3$, unconstrained}
 \end{subfigure}
 \begin{subfigure}{0.49\textwidth}
 \scalebox{0.55}{\input{figs/IN_fig_POL_C_neg_3}}
 \caption{\label{fig:covarpos_10}$d=3$, constrained} 
 \end{subfigure}
  \caption{\label{fig:covarpos}Positivity error.  The panels show the percentage
of second-moment matrices over  $n_{\test}=5{,}000$ samples that fail to be
positive semidefinite for the  joint distribution learner (JDL), the polynomial
joint distribution learner (JPDL), and the conditional mean embedding (CME) on 
the $y$-axis. The $x$-axis shows the number $n$ of data points used for training
and validation. The data are generated from a mean-zero, unit standard deviation
multivariate Gaussian distribution with covariance matrix sampled from the
algorithm proposed by \citet{ilyahensen21}. For $n\leq 10^5$, low-rank algorithm
\ref{algo:bioChol} is applied to the kernel matrices with tolerance
$\varepsilon=10^{-3}$, for $n= 10^6$ with $\varepsilon=10^{-2}$, and for
$n=10^7$ with $\varepsilon=10^{-1}$. JPDL features $\varepsilon=0$.}
\end{figure}

Next, we estimate for each simulation run $j=1,\ldots,100$ the probability that
the estimated moment matrix fails to be positive semidefinite up to machine
precision,
\[  \frac{1}{n_{\test}}\sum _{t=1}^{n_{\test}} \bigg(1-1\!\!1
 \Big(\mathbb E^{\mathbb Q_{Y|X=x^{\test,j}_t}}_{\bs \theta _j, \Mcal}[\bs
   \Tcal(Y)]\succeq 0\Big)\bigg),\]
where $1\!\!1$ denotes the indicator function and, as before, $\Mcal$ is a
placeholder for the model under consideration. Figure \ref{fig:covarpos} shows
that all methods but CME in the three-dimensional case perform well with respect
to this criterion. Similarly to the results on losses and normalization, CME
encounters problems  when the relative tolerance $\varepsilon$ of the low-rank
approximation becomes too large. Figure \ref{fig:covarpossmall}
shows positivity and normalization errors for uniformly distributed small data
sets with $n=50,100,200$. Here, JPDL shows an increased occurence of positivity
violations, in particular for $n=50$, while JDL outperforms CME, which however
only for $n=50$ suggests a probability of 5\% of  generating moment matrices
that are not symmetric positive semidefinite, and zero for more data points.
Normalization errors of CME are smaller than those generated by Gaussian data.
For heavy-tailed data generated from a Cauchy distribution, Figure \ref{fig:covarpossmall2} shows that the positivity errors are increased  in general for $n=50,100,200$. In addition to the positivity of covariance matrices,  Panel \ref{fig:covarpos_10_small2} shows the percentage of negative realizations of the model-implied Radon-Nikodym derivative. Both panels indicate that JDL exhibits the smallest violations by far.

\begin{figure}[htb]
 \begin{subfigure}{0.49\textwidth}
 \scalebox{0.55}{\input{figs/IN_fig_POL_neg_add}}
 \caption{\label{fig:uccovarpos_10_small}$d=3$, positivity,  unconstrained}
 \end{subfigure}
 \begin{subfigure}{0.49\textwidth}
 \scalebox{0.55}{\input{figs/IN_fig_POL_dev_add}}
 \caption{\label{fig:covarpos_10_small}$d=3$, normalization, unconstrained}
 \end{subfigure}
  \caption{\label{fig:covarpossmall}Positivity and normalization error for small
uniform data sets.  The panels show the percentage of second-moment matrices
over $n_{\test}=5{,}000$ samples that fail to be positive semidefinite for the
joint distribution learner (JDL), the polynomial joint distribution learner
(JPDL), and the conditional mean embedding (CME)  on the y-axis. The $x$-axis
shows the number $n$ of data points used for training and validation. The data
are generated as $(U_1, (U_1+U_2)/2,\ldots , (U_1+\cdots + U_6)/6)$, where
$U_1,\ldots,U_6$ are i.i.d. draws from the uniform distribution on the unit interval. Low-rank Algorithm \ref{algo:bioChol} is applied to the
kernel matrices with tolerance $\varepsilon=10^{-6}$. JPDL features
$\varepsilon=0$.}
\end{figure}

\begin{figure}[htb]
 \begin{subfigure}{0.49\textwidth}
 \scalebox{0.55}{\input{figs/IN_fig_POL_neg_addr2}}
 \caption{\label{fig:uccovarpos_10_small2}$d=3$, positivity,  unconstrained}
 \end{subfigure}
 \begin{subfigure}{0.49\textwidth}
 \scalebox{0.55}{\input{figs/IN_fig_POL_neg_addr22}}
 \caption{\label{fig:covarpos_10_small2}$d=3$, positivity of RN derivative, unconstrained}
 \end{subfigure}
  \caption{\label{fig:covarpossmall2}Positivity error for small
Cauchy data sets.  The left panel shows the percentage of second-moment matrices
that fail to be positive semidefinite, the right panel the percentage of a negative Radon-Nikodym derivative over $n_{\test}=5{,}000$ samples for the
joint distribution learner (JDL), the polynomial joint distribution learner
(JPDL), and the conditional mean embedding (CME, evaluated against indicator functions as $\bm I \bs F {\bs\Phi}_X^\T(\cdot)$ in \eqref{CMEeq1hat})  on the y-axis. The $x$-axis
shows the number $n$ of data points used for training and validation. The data
are generated as $(C_1, (C_1+C_2)/2,\ldots , (C_1+\cdots + C_6)/6)$, where
$C_1,\ldots,C_6$ are i.i.d. draws from the Cauchy distribution with location zero and scale 0.1. Low-rank Algorithm \ref{algo:bioChol} is applied to the
kernel matrices with tolerance $\varepsilon=10^{-6}$. JPDL features 
$\varepsilon=0$.}
\end{figure}

\begin{figure}[htb]
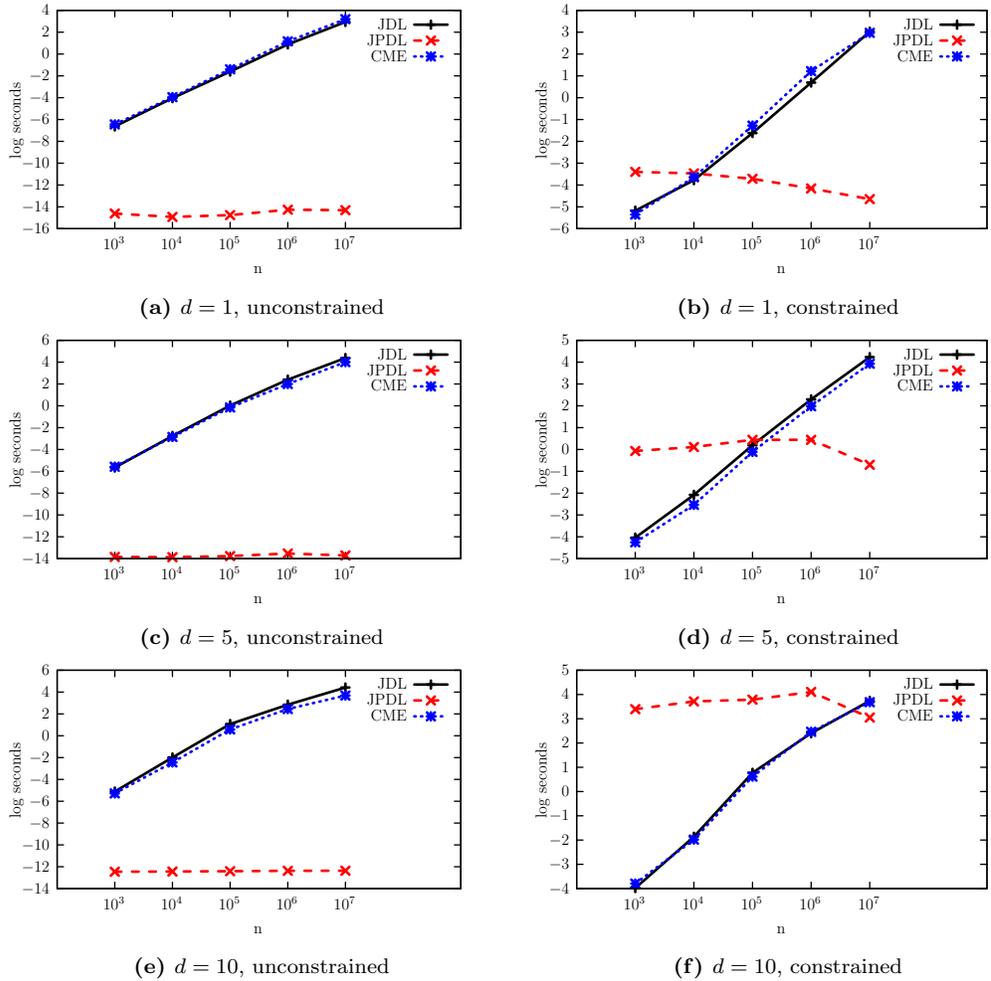

 \begin{subfigure}{0.49\textwidth}
 \scalebox{0.55}{\input{figs/IN_fig_POL_time_3}}
 \caption{\label{fig:uccovarspeed_10}$d=10$, unconstrained}
 \end{subfigure}
  \begin{subfigure}{0.49\textwidth}
 \scalebox{0.55}{\input{figs/IN_fig_POL_C_time_3}}
 \caption{\label{fig:covarspeed_10}$d=10$, constrained}
 \end{subfigure}
 \caption{\label{fig:covarspeed}Second-moment computation speed. The panels show
the running time of solving for the optimal coefficients, and  calculating
conditional expectations using  $n_{\test}=5{,}000$ samples for the joint
distribution learner (JDL), the polynomial joint distribution learner (JPDL),
and the conditional mean embedding (CME) on the $y$-axis. The $x$-axis shows the
number $n$ of data points used for training and validation. The data are
generated from a mean-zero, unit standard deviation multivariate Gaussian
distribution with covariance matrix sampled from the algorithm proposed by
\citet{ilyahensen21}. For $n\leq 10^5$, low-rank algorithm \ref{algo:bioChol} is
applied to the kernel matrices with tolerance $\varepsilon=10^{-3}$, for
$n= 10^6$ with $\varepsilon=10^{-2}$, and for $n=10^7$ with 
$\varepsilon=10^{-1}$. JPDL features  $\varepsilon=0$.}
\end{figure}

Figure \ref{fig:covarspeed} compares the methods with respect to their
computational speeds, which is measured as the time needed to solve for the
optimal coefficients, and to evaluate $n_{\test}=5{,}000$ conditional
expectations. It shows that in the unconstrained case, JDL and CME are not
distinguishable from each other. JPDL is orders of magnitudes faster in the
unconstrained case compared to JDL and CME. The unconstrained solution, which is
available in closed form, is only minimally faster than the constrained version.
The exponentially growing dimension of the subspace to describe the s.o.s. cone
\eqref{eq:semdefcone} in $d$ of JPDL becomes a significant computational burden
for JPDL for higher-dimensional problems. For instance, in the case $d=3$ and
$q=4$, the cone is described through a symmetric positive semidefinite matrix
of dimension $\binom{6+4}{6}=210$. As a consequence, JPDL outperforms the other
methods only for large sample sizes~$n$.

\subsection{Binary classification}
To assess the competing methods in a real-word  application, we consider in this
section a binary classification problem using portfolio excess return data from
\href{https://mba.tuck.dartmouth.edu/pages/faculty/ken.french/data_library.html
}{Fama-French}. The data set contains more than $25{,}000$ daily excess returns
$z_t=[z_{t,1},\dots,z_{t,25}]$ of $25$ portfolios. We are interested in
predicting the conditional probability that the excess return aggregated over
the first $d\le 25$ portfolios $y_t\isdef\sum_{i=1}^d z_{t,i}\in\Ycal=\R$ falls
below a given threshold $q$, given preceding portfolio returns 
$x_t\isdef [z_{t-1,1},\dots,z_{t-1,d}]\in\Xcal=\R^d$. In other words, the
probability that the indicator $\Ical_q(y_t)\isdef 1\! \!1(y_t\le q)$ equals 1.

Similarly to the previous experiment, we subsample $j=1,\ldots,100$ data sets 
$z_1^{(j)},\ldots, z_n^{(j)}$ of sizes $n=1{,}000,\, 5{,}000,\,10{,}000,\, 
20{,}000$, of varying numbers of portfolios $d\in\{1,5,10,25\}$. We use 
$n_{\train}=4n/5$ data points for training and the remaining $n/5$ to  validate
the kernel and regularization hyperparameters. To test the models
out-of-sample, we draw for each of the $j=1,\ldots, 100$ samples additional
$5{,}000$ data points without replacement from the remaining data set, which are
not used for training or validation.

For each experiment $(j,n,d)$, we set $q=q(j,n,d)$ to be the $1\%$-quantile of
the empirical distribution of the aggregated portfolio excess returns
$y_t^{(j)}=\sum_{i=1}^d z_{t,i}^{(j)}$, $t=1,\dots,n$.

We use the logistic loss function to validate and test all models. The logistic
loss function (and others), require normalized and positive probabilities from
the classifier. Since some of the models tested are neither pointwise
normalized nor positive, we adapt the  logistic loss function, here written for
the test data, as
  \begin{multline}\label{eq:logisticloss}
 -\frac{1}{n_{\test}}\sum _{t=1}^{n_{\test}} \Ical_q(y^{\test,j}_{t}) 
 \log \min \left(1,\max \left(\mathbb E^{\mathbb 
 Q_{Y|X=x^{\test,j}_{t}}}_{\bs \theta _j, \Mcal}[\Ical _q( Y )], 
\eta  \right)\right)\\
-\frac{1}{n_{\test}}\sum _{t=1}^{n_{\test}}(1- \Ical_q(y^{\test,j}_{t})) 
\log \max \left(\eta,\min \left(1-\mathbb E^{\mathbb 
Q_{Y|X=x^{\test,j}_{t}}}_{\bs \theta _j, \Mcal}[\Ical _q( Y )], 1 \right) 
\right),
 \end{multline}
with $\eta=10^{-12}$, where $(x^{\test,j}_{1},y^{\test,j}_{1})\dots, 
(x^{\test,j}_{n_{\test}},y^{\test,j}_{n_{\test}})$ denotes the test sample,
and $\bs \theta _j$ the validated hyperparameters pertaining to model $\Mcal$,
in experiment $j$.

\begin{figure}[htb]
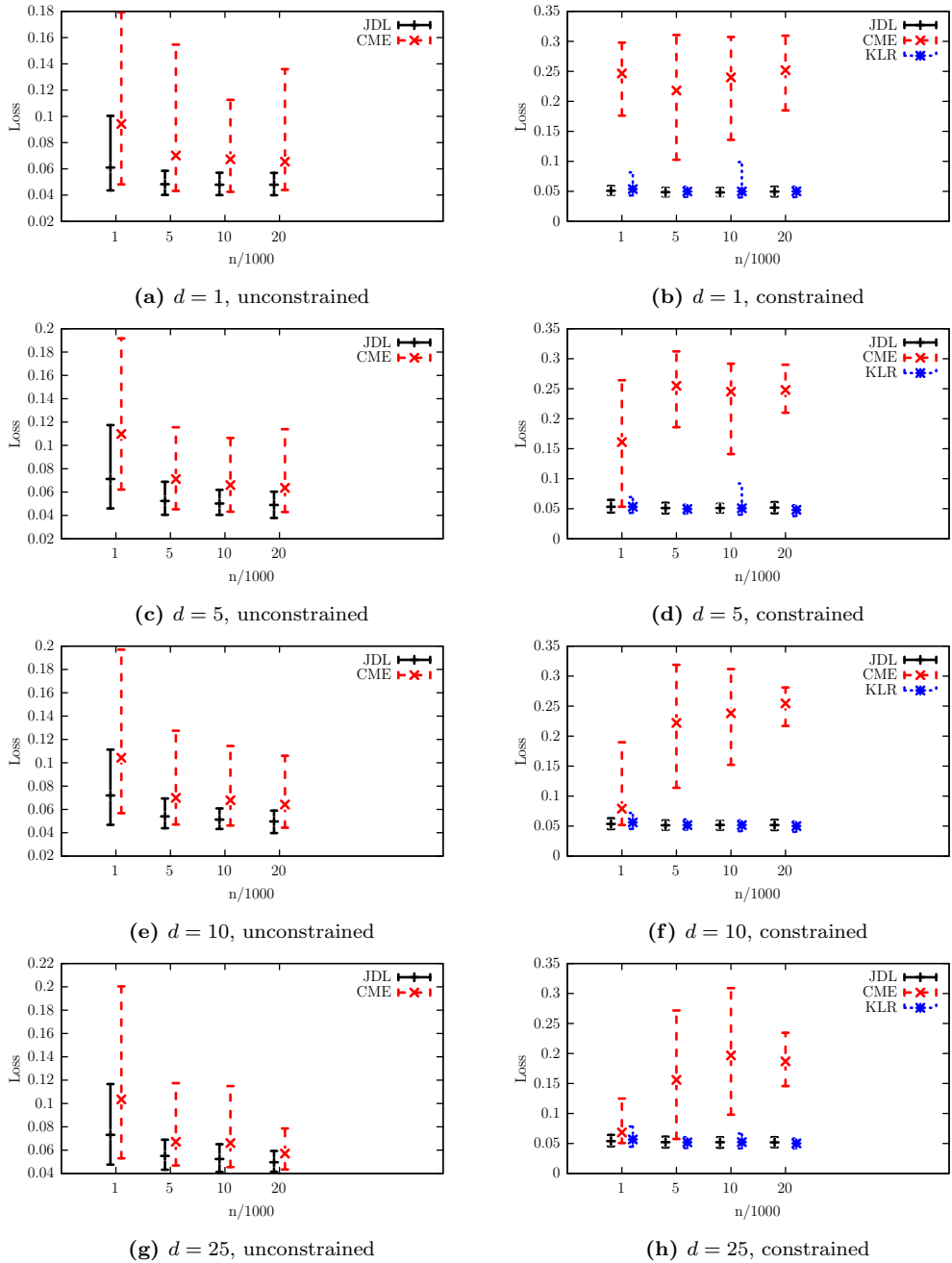

 \begin{subfigure}{0.49\textwidth}
 \scalebox{0.55}{\input{figs/IN_fig_CLASS_4}}
 \caption{\label{fig:ucclassloss_25}$d=25$, unconstrained}
 \end{subfigure}
 \begin{subfigure}{0.49\textwidth}
 \scalebox{0.55}{\input{figs/IN_fig_CLASS_C_4}}
 \caption{\label{fig:classloss_25}$d=25$, constrained}
 \end{subfigure}
  \caption{\label{fig:classloss}Logistic binary classification loss. The panels
show loss function \eqref{eq:logisticloss} evaluated over $n_{test}=5{,}000$
samples for the  joint distribution learner (JDL),  the conditional mean
embedding (CME), and logistic kernel regression (KLR) on the $y$-axis. The 
$x$-axis shows the number $n$ of data points used for training and validation.
The data are daily  portfolio returns from 
\href{
https://mba.tuck.dartmouth.edu/pages/faculty/ken.french/data_library.html
}{Fama-French} from 1927-2022.}
\end{figure}

\begin{figure}[htb]
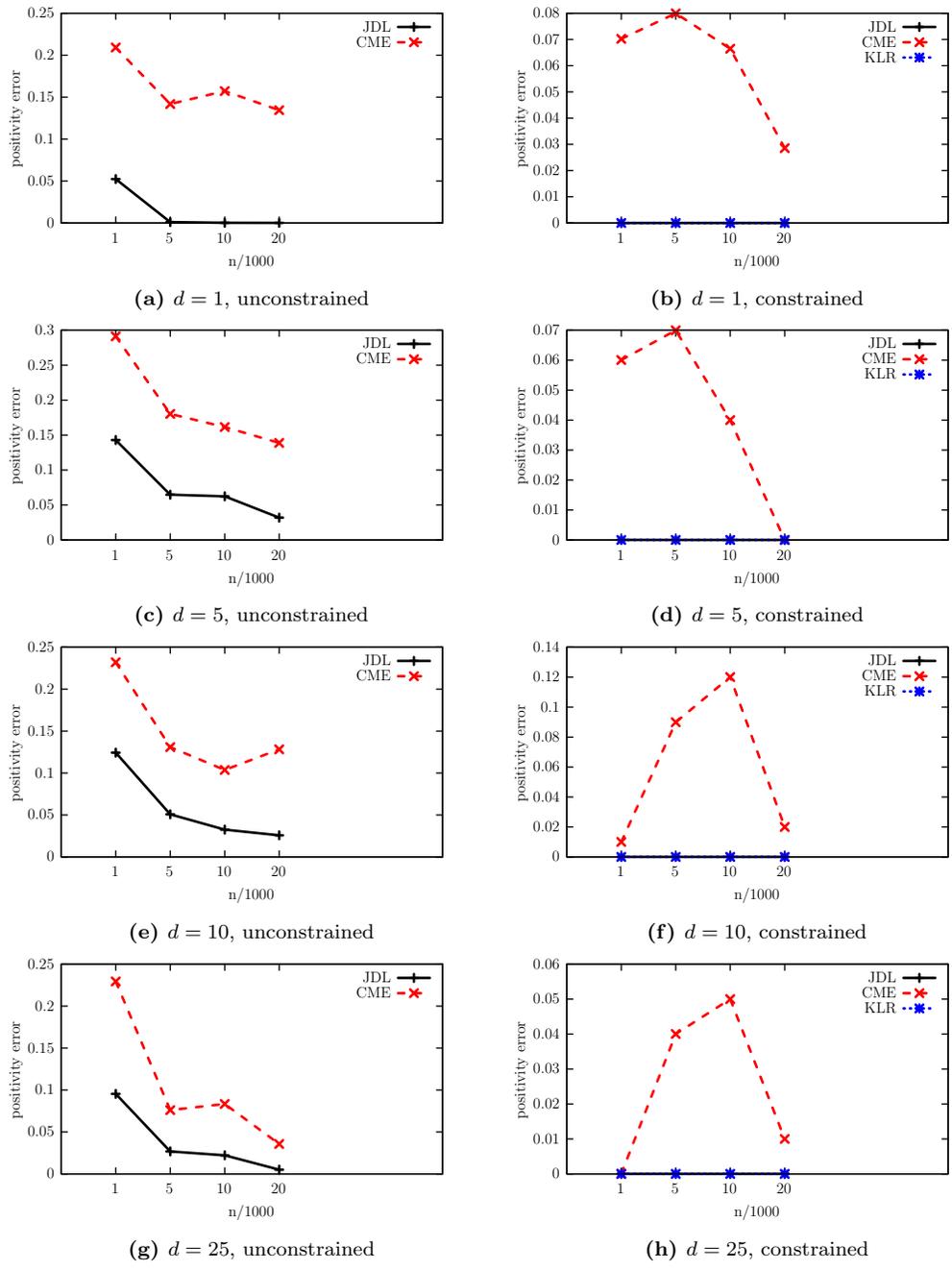

 \begin{subfigure}{0.49\textwidth}
 \scalebox{0.55}{\input{figs/IN_fig_CLASS_neg_4}}
 \caption{\label{fig:ucclasspos_25}$d=25$, unconstrained}
 \end{subfigure}
 \begin{subfigure}{0.49\textwidth}
 \scalebox{0.55}{\input{figs/IN_fig_CLASS_C_neg_4}}
 \caption{\label{fig:classpos_25}$d=25$, constrained}
 \end{subfigure}
  \caption{\label{fig:classpos}Logistic binary classification positivity error.
  The panels show the percentage of negative ``probablities'' over
$n_{test}=5{,}000$ samples for the joint distribution learner (JDL), the
conditional mean embedding (CME), and logistic kernel regression (KLR)  on the
$y$-axis. The $x$-axis shows the number $n$ of data points used for training
and validation. The data are daily  portfolio returns from 
\href{https://mba.tuck.dartmouth.edu/pages/faculty/ken.french/data_library.html
}{Fama-French} from 1927-2022.}
\end{figure}

\begin{figure}[htb]
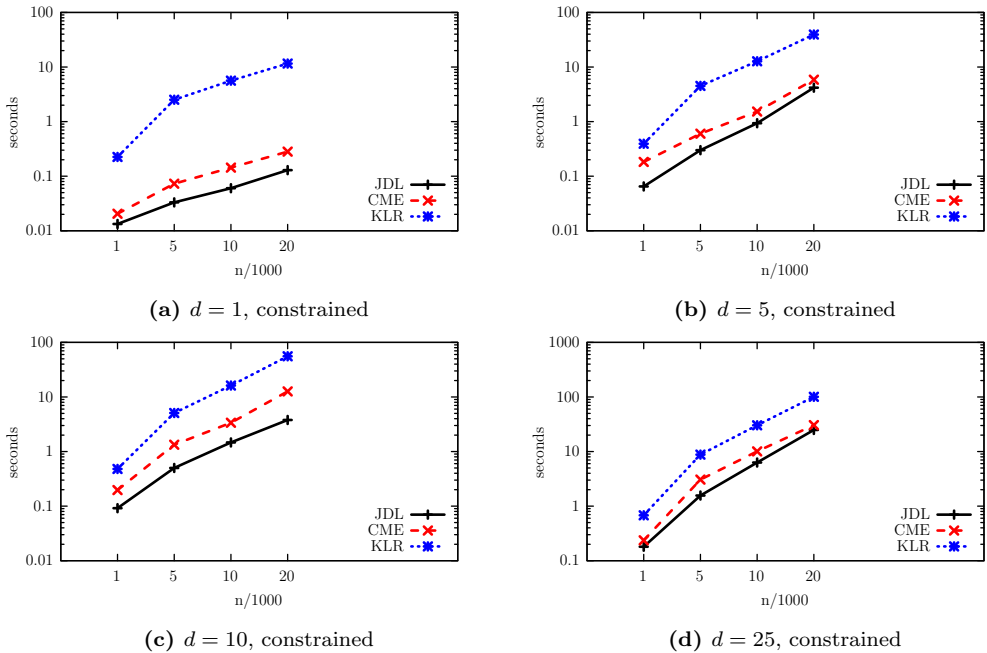

 \begin{subfigure}{0.49\textwidth}
 \scalebox{0.55}{\input{figs/IN_fig_CLASS_C_time_3}}
 \caption{$d=10$, constrained}
 \end{subfigure}
 \begin{subfigure}{0.49\textwidth}
 \scalebox{0.55}{\input{figs/IN_fig_CLASS_C_time_4}}
 \caption{$d=25$, constrained}
 \end{subfigure}
 \caption{\label{fig:classspeed}Logistic binary classification computation
 speed. The panels show the running time of calculating conditional expectations
using  $n_{test}=5{,}000$ samples for the   joint distribution learner (JDL),
the conditional mean embedding (CME), and logistic kernel regression (KLR) on
the $y$-axis. The $x$-axis shows the number $n$ of data points used for training
and validation. The data are daily  portfolio returns from 
\href{https://mba.tuck.dartmouth.edu/pages/faculty/ken.french/data_library.html
}{Fama-French} from 1927-2022.}
\end{figure}

For the JPDL, up to 26 dimensions exceed our computational means. We substitute
it with a kernel logistic regression (KLR) \citet{zhuhastie05} that is
tailor-made for the task at hand, see Appendix \ref{sec:klr}.

Figure \ref{fig:classloss} shows that, as for the conditional second-moment
matrices, JDL dominates CME for small data sizes in the unconstrained case. 
In the constrained case, JDL is on par with KLR for all experiments but small
sample sizes in high dimensions. The fit of CME is however severely hindered by
the structural constraints.

KLR produces positive classification probabilities by construction. Figure
\ref{fig:classpos} shows the estimated  likelihood of generating negative
classification probabilities for the other methods,
\[  \frac{1}{n_{\test}}\sum _{t=1}^{n_{\test}}
 \mathbbm{1}\Big(\mathbb E^{\mathbb 
   Q_{Y|X=x^{\test,j}_{t}}}_{\bs \theta _j, \Mcal}[
   \Ical
   _q( Y )]< 0\Big).\]
The structural constraints uniformly decrease the likelihood of negative
probabilities in binary classification for all the methods, to zero for JDL, and
two-fold to five-fold improvements for CME.

Figure \ref{fig:classspeed} shows that CME and JDL are comparable in speed for
the constrained case. KLR is slightly slower, but benefits greatly from its
low-rank implementation. JDL thus classifies similarly well as the specialized
KLR, but at greater speed.

\section{Conclusion}\label{sec:conclusion}
We have proposed a new approach to estimate joint and conditional distributions
in a tensor product RKHS employing adaptive low-rank approximation techniques.
Our framework is well equipped to handle large samples, while satisfying
positivity and normalization as important structural traits induced by
probability distributions. We have shown that the presented approach is
computationally efficient, versatile and successful in prediction and
classification.
\clearpage
\begin{appendix}
\section{Proofs}\label{sec:proofs}
\begin{proof}[Proof of Lemma \ref{lem:intro}] We prove the first claim only,
since the second follows similarly. In view of \eqref{QXYeq}, we have
\begin{align*}
\Q_X &= \frac{1}{n^2} \sum_{i,j=1}^n\delta_{x_i}+
    \frac{1}{n^2} \sum_{i,j=1}^nh(x_i,y_j)\delta_{x_i}=\hat\Pa_X
    +\frac{1}{n^2}\sum_{i=1}^n\bigg(\sum_{j=1}^nh(x_i,y_j)\bigg)\delta_{x_i}.
\end{align*}
Hence $\Q_X=\hat\Pa_X$ is equivalent to $\frac{1}{n^2}\sum_{i=1}^n\left(\sum_{j=1}^nh(x_i,y_j)\right)\delta_{x_i}=0$. This again holds if and only if $\sum_{i': x_{i'}=x_i}\sum_{j=1}^nh(x_{i'},y_j)=0$,  for $i = 1,\ldots, n$, as claimed.
\end{proof}
\begin{proof}[Proof of Lemma \ref{lemdist}]
Every function $f$ defined on $\Zcal$ with $\sup_{z\in\Zcal}|f(z)|\le 1$ 
satisfies $\|f\|_{L^2_{\hat \Pa_X\otimes \hat \Pa_Y}}\le 1$ by the
Cauchy--Schwarz inequality, and hence $\beta(\Q(h),\hat\Pa)\le \Ecal(h)^{1/2}$. As $\beta$ is a metric, we have $\beta(\Q(h),\Pa)\le
\beta(\Q(h),\hat\Pa)+\beta(\hat\Pa,\Pa)$, which proves the claim.
\end{proof}

\begin{proof}[Proof of Theorem \ref{thm:representer}]
First, note that for any function $h\in\Hcal$ the cost function $\Ecal(h)$ and
the linear constraints \eqref{norm} and \eqref{posi} only depend on their values
on the grid $h|_{\Gcal}$. Denote by
${\bs\Phi}_Y^\T{\bs\Phi}_X\colon\Hcal\to\R^{n\times n}$ the linear
operator $h\mapsto h|_\Gcal$, which evaluates \(h\) at \(\Gcal\). Hence, the
image of the adjoint $({\bs\Phi}_Y^\T{\bs\Phi}_X)^\ast$ consists of all 
functions of the form \eqref{eq:state_ij_id}, and it is well known that its
orthogonal complement in $\Hcal$ is the kernel of ${\bs\Phi}_Y^\T{\bs\Phi}_X$.
Now, let $h$ be any feasible function of the form \eqref{eq:state_ij_id} such
that it satisfies \eqref{norm} and \eqref{posi}. Let $f$ be any non-zero
function that is orthogonal to $h$ in $\Hcal$, and hence in the kernel of
${\bs\Phi}_Y^\T{\bs\Phi}_X$. That is, $f|_{\Gcal}=\bm 0$ element-wise on the
grid. Then $h+f$ also satisfies \eqref{norm} and \eqref{posi}, and
$\Ecal(h+f)=\Ecal(h)$, while the regularized cost function is strictly greater,
\[
\Ecal(h+f)+\lambda\|h+f\|_\Hcal^2 = \Ecal(h)+\lambda\|h+f\|_\Hcal^2>\Ecal(h)
+\lambda\|h\|_\Hcal^2,
\]
because $\lambda>0$. We conclude that the optimal $h$ is of the form
\eqref{eq:state_ij_id}.

Expressions \eqref{eq:norma} and \eqref{posiC} follow by inspection. Hence
\eqref{optgen} is equivalent to the $n^2$-dimensional problem of minimizing
$\Ecal\big(h(\bm H)\big)+\lambda \trace(\bm H^\T \bm K_Y \bm H \bm K_X)$ over
$\bm H\in \R^{n\times n}$ subject to \eqref{eq:norma} and \eqref{posiC}. It
remains to observe that this problem is equivalent to \eqref{optgenC}. Indeed,
this follows because
\begin{align*}
 &\Ecal\big(h(\bm H)\big)+\lambda \trace( \bm H^\T \bm K_Y \bm H \bm K_X) \\
 &\qquad=\frac{1}{n^2}\trace{[(n\bs I-\bs E-\bs K_Y\bs H \bs K_X)^{\T}
 (n\bs I-\bs E-\bs K_Y\bs H \bs K_X)]}+\lambda \trace{(\bs H^{\T}
 \bs K_Y\bs H \bs K_X)} \\
 &\qquad= \Rcal_\lambda(\bm H) + C,
\end{align*}
for a constant $C$ that does not depend on $\bm H$. This completes the proof of
the theorem.
\end{proof}

\begin{proof}[Proof of Theorem \ref{thmChol}]
Without loss of generality, we assume that \(p_m=m\)
for \(m\leq\rank{\bs K}\).
Otherwise, perform the proof with permuted matrix
\({\bs P}{\bs K}{\bs P}^\intercal\) with
\({\bs P}=[{\bs e}_{p_1},\ldots,{\bs e}_{p_n}]^\T\) with
\(\{p_{m+1},\ldots p_n\}=\{1,\ldots,n\}\setminus\{p_1,\ldots,p_m\}\).
The positivity of the Schur complement
$\bm K - \bm L_m \bm L_m^\T$ and the error bound \eqref{Bprop0} are well
known, see, for example, \citet{HPS12}.

To prove the other claims, we observe that \({\bs L}_m\) is a lower-triangular
matrix. From line 5 in Algorithm \ref{algo:bioChol}, it follows
that \({\bs b}_i\in\operatorname{span}\{{\bs e}_1,\ldots,{\bs e}_i\}\) for
\(i\leq m\), i.e., \({\bs B}_m\) is upper triangular. We prove inductively that
\({\bs B}_m^\T{\bs L}_m={\bs I}_{m\times m}\). For \(m=1\), there holds
\[
{\bs B}_1^\T{\bs L}_1 = {\bs b}_1^\T{\bs\ell}_1
=\frac{1}{\sqrt{d_{0,1}}}{\bs e}_1^\T{\bs\ell}_1=[1]={\bs I}_{1\times 1},
\]
since \(\ell_{1,1}=\sqrt{d_{0,1}}\). Now, let the induction hypothesis hold for
\(m-1\) and consider the block matrix
\begin{equation}\label{eq:blockInverse}
{\bs B}_m^\T{\bs L}_m =
[{\bs B}_{m-1},\bs b_m]^\T[{\bs L}_{m-1},\bs\ell_m]=
\begin{bmatrix}{\bs B}_{m-1}^\T{\bs L}_{m-1} & {\bs B}_{m-1}^\T
{\bs\ell}_m\\
{\bs b}_m^\T {\bs L}_{m-1} & {\bs b}_m^\T{\bs\ell}_m\end{bmatrix}
\end{equation}
By the induction hypothesis, there holds
\({\bs B}_{m-1}^\T{\bs L}_{m-1}={\bs I}_{(m-1)\times(m-1)}\).
Thus, since \({\bs B}_m\) is upper triangular, we obtain
\begin{equation}\label{eq:ZeroColumn}
{\bs B}_{m-1}^\T{\bs\ell}_{m}={\bs 0}\in\R^{(m-1)\times 1},
\end{equation}
and it remains to show that
\({\bs b}_m^\T{\bs L}_{m}={\bs b}_m^\T[{\bs L}_{m-1},{\bs\ell}_m]
=[0,\ldots,0,1]\in\R^{1\times m}\).
From
the induction hypothesis and \eqref{eq:ZeroColumn} it follows
that \({\bs B}_{m-1}^\T{\bs L}_{m}=[{\bs I}_{(m-1)\times(m-1)},{\bs 0}]\).
Hence, we infer
\begin{align*}
{\bs b}_m^\T{\bs L}_m&=
\frac{1}{\sqrt{d_{m-1,m}}}\big({\bs e}_m-{\bs B}_{m-1}{\bs L}_{m-1}^\T
{\bs e}_m\big)^\T{\bs L}_m=
\frac{1}{\sqrt{d_{m-1,m}}}\big({\bs e}_m^\T{\bs L}_m -{\bs e}_m^\T
{\bs L}_{m-1}
{\bs B}_{m-1}^\T{\bs L}_{m}\big)\\
&=\frac{1}{\sqrt{d_{m-1,m}}}{\bs e}_m^\T\big({\bs L}_m -{\bs L}_{m-1}
[{\bs I}_{(m-1)\times(m-1)},{\bs 0}]\big)=\frac{1}{\sqrt{d_{m-1,m}}}{\bs e}_m^\T
[{\bs 0}, {\bs\ell}_m].
\end{align*}
In view of
\(\ell_{m,m}=\sqrt{d_{m-1,m}}\), we arrive at
\[
{\bs b}_m^\T{\bs L}_m=\frac{1}{\sqrt{d_{m-1,m}}}{\bs e}_m^\T[{\bs 0}, {\bs\ell}_m]
=[0,\ldots,0,1]\in\R^{1\times m}.
\]
Inserting this into \eqref{eq:blockInverse} proves
\({\bs B}_m^\T{\bs L}_m={\bs I}_{m\times m}\).

With \(r\isdef\rank{\bs K}\), we finally have
\[
{\bs K}{\bs B}_m={\bs L}_r {\bs L}_r^\T{\bs B}_m
={\bs L}_r \begin{bmatrix}{\bs I}_{m\times m}\\ {\bs 0}
\end{bmatrix}={\bs L}_m
\quad\text{for all }m\leq r.
\]
This completes the proof.
\end{proof}
\begin{proof}[Proof of Lemma \ref{lem:positive}]
The claim follows from the elementary identity
\begin{equation}\label{elemineq}
 \min_{x\in [a,b]} x y = a y^+ - b y^-
\end{equation}
for all $y\in\R$, for any real $a\le b$.
\end{proof}

\begin{proof}[Proof of Lemma \ref{lemposi}]
First, observe that the following elementary identities hold
\begin{align}
\underline{\bs C}_{i,j} &= \min_{s,t\in\{1,\dots,n\}} [\bs L_Y\bs V_Y]_{s,i}
[\bs L_X\bs V_X]_{t,j},\label{minCeq}\\
\overline{\bs C}_{i,j} &= \max_{s,t\in\{1,\dots,n\}} [\bs L_Y\bs V_Y]_{s,i}
[\bs L_X\bs V_X]_{t,j}.\label{maxCeq}
\end{align}
Indeed, for any index pair $i,j$ at most one of the three summands on the right hand side of \eqref{minCdef} is different from zero, and represents the minimum if so. The same holds for the right hand side of \eqref{maxCdef} with the maximum. This proves \eqref{minCeq} and \eqref{maxCeq}.

Hence, for any index pair $s,t\in\{1,\dots,n\}$ we have
\begin{align*}
  (\bs L_Y\bs V_Y\tilde {\bs H} \bs V_X^{\T}\bs L_X^{\T})_{s,t} &= \sum_{i=1}^{m_Y}\sum_{j=1}^{m_X}  [\bs L_Y\bs V_Y]_{s,i} \tilde H_{i,j} [\bs L_X\bs V_X]_{t,j}\\
  &\ge \sum_{i=1}^{m_Y}\sum_{j=1}^{m_X} \left( \underline{\bs C}_{i,j} {\tilde H_{i,j}}^+ - \overline{\bs C}_{i,j} {\tilde H_{i,j}}^-\right)
  = \trace\big(\underline{\bs C}^\T \tilde {\bs H}^+\big)-\trace\big(\overline{\bs C}^\T \tilde {\bs H}^-\big),
\end{align*}
where we used \eqref{elemineq}. This proves the claim.
\end{proof}
\begin{proof}[Proof of Lemma \ref{lem:approx}]
The expression on the right hand side of \eqref{eq:orthoproj} follows from an application of the matrix derivative. Furthermore, by orthogonality of $h-h_{\tilde\Hcal}$ and $h_{\tilde\Hcal}-\tilde h$, we have
\begin{align*}
 \|h-\tilde h\|_{\Hcal}^2 =\| h-h_{\tilde\Hcal}\|_{\Hcal}^2 + \|h_{\tilde\Hcal}-\tilde h\|_{\Hcal}^2 =\| h-h_{\tilde\Hcal}\|_{\Hcal}^2 + \| \bm H_{\tilde\Hcal}-\tilde{\bm H} \|_F^2 .
\end{align*}
The second equality follows from the orthonormality of $\bm \Psi _X(\cdot)$ in $\Hcal _\Xcal$ and $\bm \Psi _Y(\cdot)$ in $\Hcal _\Ycal$, see \eqref{orthoPsiH}. It remains to bound the first term on the right hand side from above:
\begin{align*}\|h-h_{\tilde\Hcal}\|_\Hcal^2&=
\big\|{\bs\Phi}_Y(\cdot)\bm H {\bs\Phi}_X(\cdot)^\T-{\bs\Psi}_Y(\cdot)\bm H_{\tilde\Hcal} {\bs\Psi}_X(\cdot)^\T \big\|_\Hcal^2\\
&=\big\|{\bs\Phi}_Y(\cdot)(\bm H -\bm B_Y\bm L_{Y}^{\T}\bm H\bm L_{X}\bm B_X^{\T}) {\bs\Phi}_X(\cdot)^\T\big\|_\Hcal^2\\
&=\trace \big ((\bm H -\bm B_Y\bm L_{Y}^{\T}\bm H\bm L_{X}\bm B_X^{\T})^{\T}\bm K_Y (\bm H -\bm B_Y\bm L_{Y}^{\T}\bm H\bm L_{X}\bm B_X^{\T})\bm K_X \big ) \\
&=\trace \big (\bm H^{\T}\bm K_Y \bm H \bm K_X \big ) -2\trace \big (\bm H ^{\T}\bm K_Y \bm B_Y\bm L_{Y}^{\T}\bm H\bm L_{X}\bm B_X^{\T} \bm K_X\big) \\
&\qquad+\trace \big (\bm B_X\bm L_{X}^{\T}\bm H^{\T}\bm L_{Y}\bm B_Y^{\T} \bm K_Y \bm B_Y\bm L_{Y}^{\T}\bm H\bm L_{X}\bm B_X^{\T} \bm K_X\big)\\
&=\trace \big (\bm H^{\T}\bm K_Y \bm H \bm K_X \big )-\trace \big (\bm H^{\T}\bm L_Y\bm L_Y^{\T} \bm H \bm L_X \bm L_X^{\T} \big )\\
&=\trace \big (\bm H^{\T}\bm K_Y \bm H (\bm K_X-\bm L_X\bm L_X^\T) \big)
+\trace \big (\bm H^{\T}(\bm K_Y-\bm L_Y\bm L_Y^{\T})\bm H \bm L_X \bm L_X^{\T} \big )\\
&=\langle\bm K_Y, \bm H (\bm K_X-\bm L_X\bm L_X^\T)\bm H^\T\rangle_F
+\langle\bm L_X \bm L_X^{\T},\bm H^\T(\bm K_Y-\bm L_Y\bm L_Y^{\T})\bm H\rangle_F
\\
&\leq\|{\bs H}\|_F^2\|{\bs K}_Y\|_F
\|\bm K_X-\bm L_X\bm L_X^\T \|_F+\|{\bs H}\|_F^2\|\bm L_X \bm L_X^{\T}\|_F
\|\bm K_Y-\bm L_Y\bm L_Y^\T \|_F\\
&\leq \varepsilon \|{\bs H}\|_F^2\big(\trace{\bs K}_X+\trace{\bs K}_Y),
\end{align*}
where we used the Cauchy--Schwarz inequality for the Frobenius inner product, as well as the sub-multiplicativity of the Frobenius norm, and the fact that \(\|{\bs A}\|_F\leq\trace{\bs A}\) for any positive semidefinite matrix $\bs A$,
and \(\trace\bm L_X \bm L_X^{\T}\leq\trace{\bs K}_X\).
\end{proof}

\section{Polynomial distributions}\label{sec:polyembedding}
Finite-dimensional RKHS lead to low-rank problems by construction, and in this
section, we specialize our approach to polynomial kernels that are particularly
amenable to positivity constraints. In this setting, the rank of the kernel
matrices is predetermined from the degree of the polynomial, and does not
necessitate additional low-rank techniques, as described in the previous section
for large data sets. Furthermore, as we will see below, the polynomial kernel
basis with a particularly suitable inner product is automatically
double-orthogonal, without requiring further modifications or rotations.

Let $\Xcal\subseteq \R^{d_X}$ and $\Ycal \subseteq \R^{d_Y}$, with non-empty
interiors, and choose a common polynomial degree $q$ for simplicity. Let
\(
\Pcal _q(\Xcal)\isdef\spn\big\{{x}^{\bs\alpha}:
\bs\alpha\in\mathbb{N}^{d_X},|\bs\alpha|\leq q\big\}\) and
\(\Pcal _q(\Ycal)\isdef\spn\{{y}^{\bs\alpha}:
\bs\alpha\in\mathbb{N}^{d_Y},|\bs\alpha|\leq q\},
\)
denote the spaces of all polynomials on $\Xcal$ and $\Ycal$ of degree \(q\) or
less. The dimensions of the vectors of monomials generating $\Pcal _q(\Xcal)$
and $\Pcal _q(\Ycal)$, 
\(\bs \tau _{X,q}( x)\isdef [\tau_{X,q,1}(x),\dots,\tau_{X,q,m_X}(x)],\)
and
\(\bs \tau _{Y,q}( y)\isdef [\tau_{Y,q,1}(y),\dots,\tau_{Y,q,m_Y}(y)]),\)
are \(m_X=\binom{d_X+q}{d_X}\) and \( m_Y=\binom{d_X+q}{d_X},
 \)
 respectively.

To use these polynomial bases in the context of loss function
\eqref{Rcaldef}, we  use expectation inner products in terms of $\hat \Pa_X$
and $\hat \Pa _Y$, respectively. Assuming that the Gram matrices
\[  \bs G_{X,q}\isdef
 \bigg[\int _{\Xcal}  \tau_{X,q,i}\, \tau_{X,q,j} \dd\hat \Pa_X
 \bigg]_{i,j=1}^n\quad\text{and}\quad
 \bs G_{Y,q}\isdef \bigg[\int _{\Ycal} \tau_{Y,q,i}\, \tau_{Y,q,j}
 \dd\hat \Pa_Y\bigg]_{i,j=1}^n,\]
have full rank, we define the kernels $k_{X,q}:\Xcal\times \Xcal
\mapsto \R$ and $k_{Y,q}:\Ycal\times \Ycal \mapsto \R$ by
\[  k_{X,q}(x,x')=\bs \tau_{X,q}(x)\bs G^{-1}_{X,q}\bs \tau^{\T}_{X,q}(x')
 \text{ and }k_{Y,q}(y,y')=\bs \tau_{Y,q}(y)\bs G^{-1}_{Y,q}\bs 
 \tau^{\T}_{Y,q}(y').\]
For details on polynomial kernels, we refer to
\citet{berlinetthomas-agnan04}. Defining the Vandermonde matrices
\[  \bs V_{X,q}\isdef \begin{bmatrix}
                \bs \tau _{X,q}( x_1) \\
                \vdots \\
                \bs \tau _{X,q}( x_n)
               \end{bmatrix}, \, \text{ and }
  \bs V_{Y,q}\isdef \begin{bmatrix}
                \bs \tau _{Y,q}( y_1) \\
                \vdots \\
                \bs \tau _{Y,q}( y_n)
               \end{bmatrix},\]
the corresponding kernel matrices have natural decompositions,
\[  \bs K_{X}= \bs V_{X,q} \, \bs G_{X,q}^{-1/2} \bs G_{X,q}^{-1/2} \,
 \bs V_{X,q}^{\T}= \bs L_{X} \bs L_{X}^{\T} \text{ and }
 \bs K_{Y}= \bs V_{Y,q} \, \bs G_{Y,q}^{-1/2} \bs G_{Y,q}^{-1/2} \,
 \bs V_{Y,q}^{\T}= \bs L_{Y} \bs L_{Y}^{\T}.\]
These decompositions correspond to the orthogonal bases
\(\bs \tau _{X,q}{\bs G}_{X,q}^{-1/2}\) and
\(\bs \tau _{Y,q}{\bs G}_{Y,q}^{-1/2}\),
respectively.
Especially, for either kernel matrix we have $\bs L_X^{\T}\bs
L_X=\bs L_Y^{\T}\bs L_Y=n\bs I$,
immediately identifying
\[\bs B_X=\bs L_X /n\text{ and } \bs B_Y=\bs L_Y /n,\]
which satisfy the requirements \eqref{Bprop0}--\eqref{Bprop1} with tolerance
$\varepsilon=0$, as well as double orthogonality in both $\Hcal$ and
$L^2_{\hat \Pa_X\otimes \hat \Pa_Y}$, without further modifications.

Having provided the necessary ingredients, and setting $\bs H =
\bs B_Y \tilde{\bs H}\bs B_X^{\T}=\frac{1}{n^2}\bs L_{Y}\tilde{\bs H}
\bs L_{X}^{\T}$, the objective function \eqref{Rcaldef} simplifies to
\begin{equation}\label{eq:polynomial}
    (1+\lambda) \trace{(\tilde {\bs H}^{\T}\tilde {\bs H})}
    -\frac{2}{n}\trace{(\bs L_Y\tilde {\bs H} \bs L_X^{\T})}
    +\frac{2}{n^2}\trace{(\bs 1^{\T}\bs L_Y\tilde {\bs H}\bs L_X^{\T}\bs 1 )}.
\end{equation}

On top of their innate  low-rank nature, polynomials grant one additional
advantage in that they lend themselves to a tightening of the positivity
\eqref{posi} on the grid $\Gcal$ to a global pointwise positivity condition.
From the tensor product space $\Hcal=\Pcal_q(\Xcal)\otimes \Pcal _q(\Ycal)
\subset\Pcal_{2q}(\Zcal)$, we define
\begin{equation}\label{eq:semdefcone}
  \Pcal_{\textnormal{sos},2 q}(\Zcal)\isdef \{ \xi \in \Pcal_{2q} (\Zcal): 
    \xi=\bs \tau_{q}\,
 \bs A \, \bs \tau _{q}^{\T}\text{ with }\bs A\succeq 0 \},
\end{equation}
the sum-of-squares cone of degree $2q$ polynomials on $\Zcal$, where $\bs\tau_q$
here denotes the basis monomials in $z$ of degree $q$ or 
less.\footnote{Sum-of-square positivity constraints are tightly related to the
moment problem \citet{lasserre10, schmuedgen17}, and have been prominent also in
machine learning \citet{muzellec2021note}.} Through the tensor
product formulation, global pointwise positivity can in the polynomial case
thus be ensured through the semidefinite constraint 
$1+h\in \Pcal_{\textnormal{sos},2q}(\Zcal)$.

\section{Conditional Mean Embedding}\label{sec:conditionaldistributionembedding}
Here, we introduce briefly conditional mean embedding as proposed by
\citet{songetal09}, who employ the same hypothesis space, a tensor product
RKHS $\Hcal=\Hcal _X \otimes \Hcal _Y$, that is used in the previous sections to
estimate, for any $x\in \Xcal$, the conditional expectation operator 
$\mu _{Y|X=x}\in \Hcal _Y$ acting as a linear functional on $\Hcal _Y$
\begin{equation}\label{CMEeq1}
 \langle f,\mu _{Y|X=x} \rangle _{\Hcal _{Y}}=\int _{\Ycal} f\dd \mathbb
 P_{Y|X=x},\quad\text{for all $f\in \Hcal _Y$.}
\end{equation}

Consider a sample $(x_i, y_i)$, $i=1,\ldots, n$ as above. Assuming that there
exists an element $\mu_{Y\mid X=\cdot}\in\Hcal$ satisfying \eqref{CMEeq1} for
all $x\in\Xcal$, \citet{gruenewaelderetal12} show that among
$F\colon\Xcal \mapsto \Hcal _Y, \, F \in \Hcal$, the best estimator is given by
\begin{equation}\label{eq:traditional}
 \hat{\mu} _{Y|X=\cdot}\isdef \argmin _{F\in \Hcal}\big\{\frac{1}{n}
 \sum _{i=1}^n\|k_{\Ycal}(y_i,\cdot)-F(x_i)\|^2 _{\Hcal _Y}
+\lambda \|F\|_{\Hcal}^2\big\}.
\end{equation}
Note, however, that the validity of this assumption is difficult to test in
practice, as discussed, for example, in \citet{klebanovschustersullivan20}.

Leaving aside these difficulties, the optimal $F$ in \eqref{eq:traditional} is
found to be of the form
\(F ={\bs\Phi}_Y(\cdot){\bs F}\,
{\bs\Phi}_X(\cdot)^\T,\)
where ${\bs\Phi}_Y(\cdot)$ and ${\bs\Phi}_Y(\cdot)$ are given as in 
\eqref{PhiXYdef}, which follows from \citet{micchellipontil05}.
With this bilinear form, solving \eqref{eq:traditional} thus yields the solution
 \begin{equation}\label{eq:conditionaldistributionembedding}
  \bm F=(\bm K_X+n\lambda \bm I_n)^{-1}, \, \text{ and  } 
  \hat{\mu} _{Y|X=\cdot}= {\bs\Phi}_Y(\cdot)(\bm K_X
  +n\lambda \bm I_n)^{-1} {\bs\Phi}_X(\cdot)^\T.
 \end{equation}
The estimator of the conditional expectation operator \eqref{CMEeq1} then reads
\begin{equation}\label{CMEeq1hat}
 \langle f,\hat{\mu}_{Y|X=x} \rangle _{\Hcal _{Y}}=\bs [f(y_1),\dots,f(y_n)] 
 \bs F {\bs\Phi}_X^\T(x),\quad\text{for all $f\in \Hcal _Y$.}
\end{equation}
From \eqref{CMEeq1hat}, it is noteworthy, that  the optimal conditional mean
embedding does not feature the kernel matrix $\bm K _Y$. The RKHS $\Hcal _Y$
appears only to evaluate functions.  However, working with large data sets
becomes prohibitive, due to the matrix inversion in
\eqref{eq:conditionaldistributionembedding}.

\subsection{Low-rank approximation of the Conditional Mean Embedding}
\label{sec:conditionaldistemblowrank}
 To facilitate low-rank approximations as in Section~\ref{sec:lowrank}, we first
 need to  develop the objective function.
 Multiplying the objective function with $n$ and omitting the term that does not
 depend on $\bm F$, we can write the objective function \eqref{eq:traditional} as
\begin{equation}\label{eq:condembobj}
\begin{aligned}
  \Rcal_\lambda^{\textnormal{CME}}(\bs F)&\isdef 
  \sum _{i=1}^n\left \{-2{\bs\Phi}_Y(y_i)\, \bm F \, {\bs\Phi}_X^{\T}(x_i)
+{\bs\Phi}_X^{\T}(x_i)\bm F ^{\T}\bm K_{Y} \, 
\bm F {\bs\Phi}_X^{\T}(x_i)\right \}\\&\quad 
+n \lambda \trace (\bm F ^{\T}\bm K_{Y} \bm F \bm K_{X}) \\
&=-2 \trace (\bm K_{Y}\bm F \bm K_{X})
+\trace ( \bm F^{\T}\bm K_{Y}\bm F \bm K_{X}\bm K_{X})
+n \lambda \trace (\bm F ^{\T}\bm K_{Y} \bm F \bm K_{X})\\
&=-2 \trace (\bm K_{Y}\bm F \bm K_{X})
+  \trace (\bm F ^{\T}\bm K_{Y} \bm F \bm K_{X}(\bm K_{ X}+n \lambda \bm I_n)) .
\end{aligned}
\end{equation}
From the first order condition of optimality with respect to the matrix $\bm F$,
the solution \eqref{eq:conditionaldistributionembedding} evolves from this
objective function in coordinate form.\footnote{\label{foot:lucky}Taking the
matrix derivative of \eqref{eq:condembobj} with respect to $\bm F$, yields the
condition
\[  -2\bm K_Y\bm K_X+2\bm K_Y \bm F (\bm K_X+n \lambda \bm I_n)\bm K_X =\bm 0,\]
from which it becomes apparent how the solution is independent of $\bm K_Y$
at the optimum.}

The proposal in \citet{gruenewaelderetal12} to sparsify conditional mean
embedding necessitates the calculation of the full kernel matrix $\bm K_X$,
which may become prohibitive for large sample sizes. To avoid calculations with
large matrices, we consider a proposal following the low-rank 
Section~\ref{sec:lowrank}. Inserting
 \begin{equation}\label{dumeq0}
     \bm F=\bm B_{Y}\bm V _Y \tilde {\bm F}\bm V _X^\T\bm B_{X}^{\T}
 \end{equation}
 into the objective function \eqref{eq:condembobj}, for 
 $\tilde {\bm F} \in \mathbb R ^{m_Y\times m_X}$, we obtain the low-rank
 objective function on $\R^{m_Y\times m_X}$

 \begin{equation}\label{eq:lowrankprog}
 \begin{aligned}
   \tilde{\Rcal}_\lambda^{\textnormal{CME}}(\tilde {\bm F}) &
   \isdef  {\Rcal}_\lambda^{\textnormal{CME}}(\bm B_{Y}\bm V _Y \tilde {\bm F}
     \bm V _X^\T\bm B_{X}^{\T})\\
     &=-2 \trace(\bm L_Y\bm V_Y\tilde {\bm F}\bm V_X^{\T}\bm L_X ^{\T})
     +\trace(\tilde {\bm F}\bm \Lambda _X\tilde {\bm F}^{\T}
     +n \lambda \tilde {\bm F}\tilde {\bm F}^{\T}) .
 \end{aligned}
 \end{equation}

To work with the optimal solution \eqref{eq:conditionaldistributionembedding}
without low-rank approximation, one never has to make explicit $\Hcal _Y$. One
rather merely assumes that it can always be specified to accommodate the
functions to be evaluated therein. In the low-rank objective function
\eqref{eq:lowrankprog} the low-rank kernel matrix  does appear, however. As a
consequence function evaluation is also performed only on the reduced subspace.
We refer to the low-rank conditional mean embedding as CME.

\subsection{Structural constraints for the Conditional Mean Embedding}
\label{sec:conditionaldistembconst}
 Similarly to the  distribution estimator proposed in this paper, and assuming 
 there exists $f\in\Hcal _Y$ with $f(y_i)=1$ for $i=1,\dots,n$, we can impose
 normalization of the conditional distributions as the $n$ vector constraint
 \begin{equation}\label{normCMEfull}
  \bm 1^{\T}\bm F \bm K_X=\bm 1^{\T}\bm B_{Y}\bm V_Y\tilde{\bm F}
  \bm V_X^{\T} \bm L_X^{\T}=\bm 1 ^{\T}.
 \end{equation}
 Likewise, assuming that there exist $f_1,\dots,f_n\in \Hcal _Y$ such that 
 $f_i(y_j)=\delta_{ij}$ for $i,j=1,\dots,n$, positivity can be imposed as the 
 $n\times n$ matrix inequality
\begin{equation}\label{posCMEnaive}
 \bm F \bm K_X=\bm B_{Y}\bm V_Y\tilde{\bm F}\bm V_X^{\T} \bm L_X^{\T}\geq  
 \bs 0_{n\times n}, \text{ element-wise.}
\end{equation}
Note that only the $m_Y$ non-zero rows $p_{Y,1},\dots,p_{Y,m_Y}$ of 
$\bm B_{Y}$ in \eqref{posCMEnaive} are relevant. Hence the effective dimension
of \eqref{posCMEnaive} is significantly smaller than $n\times n$. We therefore
define the $m_Y\times n$ matrix
\[ \tilde{\bs B}_Y\isdef \bs P_Y^{\T} \bs B_Y,\]
consisting of the $m_Y$ non-zero rows $p_{Y,1},\dots,p_{Y,m_Y}$ of $\bs B_Y$,
for the $n\times m_Y$ matrix $\bs P_Y\isdef (\bs e_{p_{Y,1}},\dots,
\bs e_{p_{Y,m_Y}})$. We then replace \eqref{posCMEnaive} by the equivalent
$m_Y\times n$ matrix condition
\begin{equation}\label{posCME}
  \tilde{\bs B}_{Y}\bm V_Y\tilde{\bm F}\bm V_X^{\T} \bm L_X^{\T}\geq
  \bs 0_{m_Y\times n}, \text{ element-wise.}
\end{equation}
Putting everything together, constrained low-rank CME solves the program
\begin{equation}\label{optgenCME}
 \underset{\tilde {\bs F }\in \R ^{m_Y\times m_X}\text{ s.t.\
 \eqref{normCMEfull}, \eqref{posCME}}}{\text{ minimize }} 
 \tilde{\Rcal}_\lambda^{\textnormal{CME}}(\tilde {\bm F}) .
\end{equation}

There are two major computational issues for the implementation of 
\eqref{optgenCME}. First, the normalization constraint \eqref{normCMEfull}
imposes $n$ linear restrictions, which will likely be infeasible, in particular
for large $n$ compared to the dimension $m_X m_Y$ of the argument
$\tilde{\bm F}$.  We therefore significantly weaken this linear constraint, and 
impose normalization only for the unconditional mean equivalent
 \begin{equation}\label{normCME}
  \bm 1^{\T}\bm B_{Y}\bm V_Y\tilde{\bm F}\bm V_X^{\T} \bm L_X^{\T}\bm 1/n = 1.
 \end{equation}
Second, for large $n$, the positivity constraint \eqref{posCME} becomes 
computationally difficult. A tightening along the lines of \eqref{eq:tightening}
turns out to be infeasible in our numerical examples. An alternative is the
relaxation of the positivity constraint \eqref{posCME}, which is similar to 
\eqref{normCME}, namely the unconditional positivity constraint
 \begin{equation}\label{posCMEave}
  \tilde{\bs B}_{Y}\bm V_Y\tilde{\bm F}\bm V_X^{\T} \bm L_X^{\T}\bs 1\geq 
  \bs 0_{m_Y}, \text{ element-wise.}
\end{equation}

\section{Kernel logistic regression}\label{sec:klr}
We employ the standard kernelized classifier for a coefficient vector 
$\bs c\in \R ^n$, $\operatorname{cl}(\cdot)\isdef
\big(1+\exp(- \bs \Phi_X (\cdot) \bs c\big)^{-1}$, and mapping
\[ x\mapsto \begin{cases}
            1, & \operatorname{cl}(x)\geq 1/2, \\
            0, & \operatorname{cl}(x) < 1/2.
          \end{cases}\]
Applying the transformation $\bs c=\bs B_X \tilde{\bs c}$, and defining 
$\tilde {\bs \Phi }_X (\cdot)\isdef \bs \Phi_X (\cdot)\bs B_X$ yields the 
low-rank classifier
\(  \tilde{\operatorname{cl}}(\cdot)= 
  \big(1+\exp(- \tilde {\bs \Phi}_X (\cdot) \tilde {\bs c}\big)^{-1}\).
With regularization, the optimal coefficients minimizing the logistic loss, as
in \eqref{eq:logisticloss}, solve
\[  \underset{\tilde{\bs c}\in \R ^{m_X}}{\text{ minimize }} 
 \bigg\{-\sum _{i=1}^n\mathcal I_q(y_i)\log \tilde{\operatorname{cl}}(x_i)
 -\sum _{i=1}^n\big(1-\mathcal I_q(y_i)\big)
 \log\big(1-\tilde{\operatorname{cl}}(x_i)\big)
+\lambda \|\tilde {\bm c}\|_2^2\bigg\},\]
for $\lambda >0$, where we have made use of bi-orthogonality in the 
regularization.
\end{appendix}

\bibliographystyle{plainnat}
\bibliography{Bibliography}

\end{document}